\documentclass[journal]{IEEEtranTIE}

\usepackage{amsmath}
\usepackage{amssymb}

\usepackage{picinpar}
\usepackage{pifont}
\usepackage{alltt}

\usepackage{ifpdf}

\usepackage{graphicx}
\ifpdf
 \usepackage{epstopdf}
\fi

\usepackage{ragged2e} 
\usepackage{siunitx} 

\usepackage{array}
\usepackage{tabularx}
\usepackage[caption=false]{subfig}

\usepackage{tikz}
\usepackage[breaklinks=true]{hyperref}
\usepackage{breakurl}

\makeatletter
\def\@oddfoot{\hbox{}\hss\@IEEEfooterstyle\raisebox{0pt}[0pt][0pt]{\@IEEEpubid}\hss\hbox{}}\relax
\def\@evenfoot{\hbox{}\hss\@IEEEfooterstyle\raisebox{0pt}[0pt][0pt]{\@IEEEpubid}\hss\hbox{}}\relax
\def\ps@IEEEtitlepagestyle{%
  \def\@oddfoot{\hbox{}\hss\@IEEEfooterstyle\raisebox{0pt}[0pt][0pt]{\@IEEEpubid}\hss\hbox{}}\relax
  \def\@evenfoot{\hbox{}\hss\@IEEEfooterstyle\raisebox{0pt}[0pt][0pt]{\@IEEEpubid}\hss\hbox{}}\relax
}
\IEEEsettopmargin{t}{54pt}
\makeatother

\newcommand{\argmin}{\operatornamewithlimits{argmin}}
\newcommand{\argmax}{\operatornamewithlimits{argmax}}

\makeatletter
\newif\if@shownames

\@shownamestrue    

\begin{document}

\def\leftmark{\begin{minipage}{\textwidth}
\fontsize{6pt}{6.5pt}\selectfont \centering 
	\makebox[0pt][c]{This article has been accepted for publication in a future issue of this journal, but has not been fully edited. Content may change prior to final publication. Citation information: DOI 10.1109/TIE.2016.2580125, IEEE}\\
	
	Transactions on Industrial Electronics\\

	\flushleft \footnotesize IEEE TRANSACTIONS ON INDUSTRIAL ELECTRONICS
\end{minipage}}

\IEEEpubid{\fontsize{6pt}{6.5pt}\selectfont \centering 0278-0046 \textcopyright 2016 IEEE. Personal use is permitted, but republication/redistribution requires IEEE permission. See http://www.ieee.org/publications\_standards/publications/rights/index.html for more information.}

\title{Controlling Robot Morphology from Incomplete Measurements}

\if@shownames
\author{
		Martin~Pecka,
		Karel~Zimmermann,~\emph{Member,~IEEE,}
        Michal~Reinstein,~\emph{Member,~IEEE,}
        Tomas~Svoboda,~\emph{Member,~IEEE,}
		
		\thanks{
			Manuscript received November 30, 2015; revised April 7, 2016 and May 10, 2016; accepted May 11, 2016.
			The research leading to these results has received funding from the European Union under grant agreement FP7-ICT-609763 TRADR; from the Czech Science Foundation under Project GA14-13876S, and by the Grant Agency of the CTU Prague under Project SGS15/081/OHK3/1T/13.
			
			All authors are with the Dept. of Cybernetics, Faculty of Electrical Engineering, Czech Technical University in Prague, Czech republic.
			K.~Zimmermann is the corresponding author (phone: +420-22435-5733, email: zimmerk@fel.cvut.cz).
			
			M.~Pecka and T.~Svoboda are partly with the Czech Institute of Cybernetics Robotics and Informatics, Czech Technical University in Prague, Czech republic.
		}
}
\fi

\maketitle

\begin{abstract}
  Mobile robots with complex morphology are essential for traversing rough terrains in Urban Search \& Rescue missions (USAR). Since teleoperation of the complex morphology causes high cognitive load of the operator, the morphology is controlled autonomously. The autonomous control measures the robot state and surrounding terrain which is usually only partially observable, and thus the data are often incomplete. We marginalize the control over the missing measurements and evaluate an explicit safety condition. If the safety condition is violated, tactile terrain exploration by the body-mounted robotic arm gathers the missing data.
\end{abstract}


\section{Introduction}

\begin{figure}[t!]
  \centering
	\includegraphics[height=0.41\linewidth]{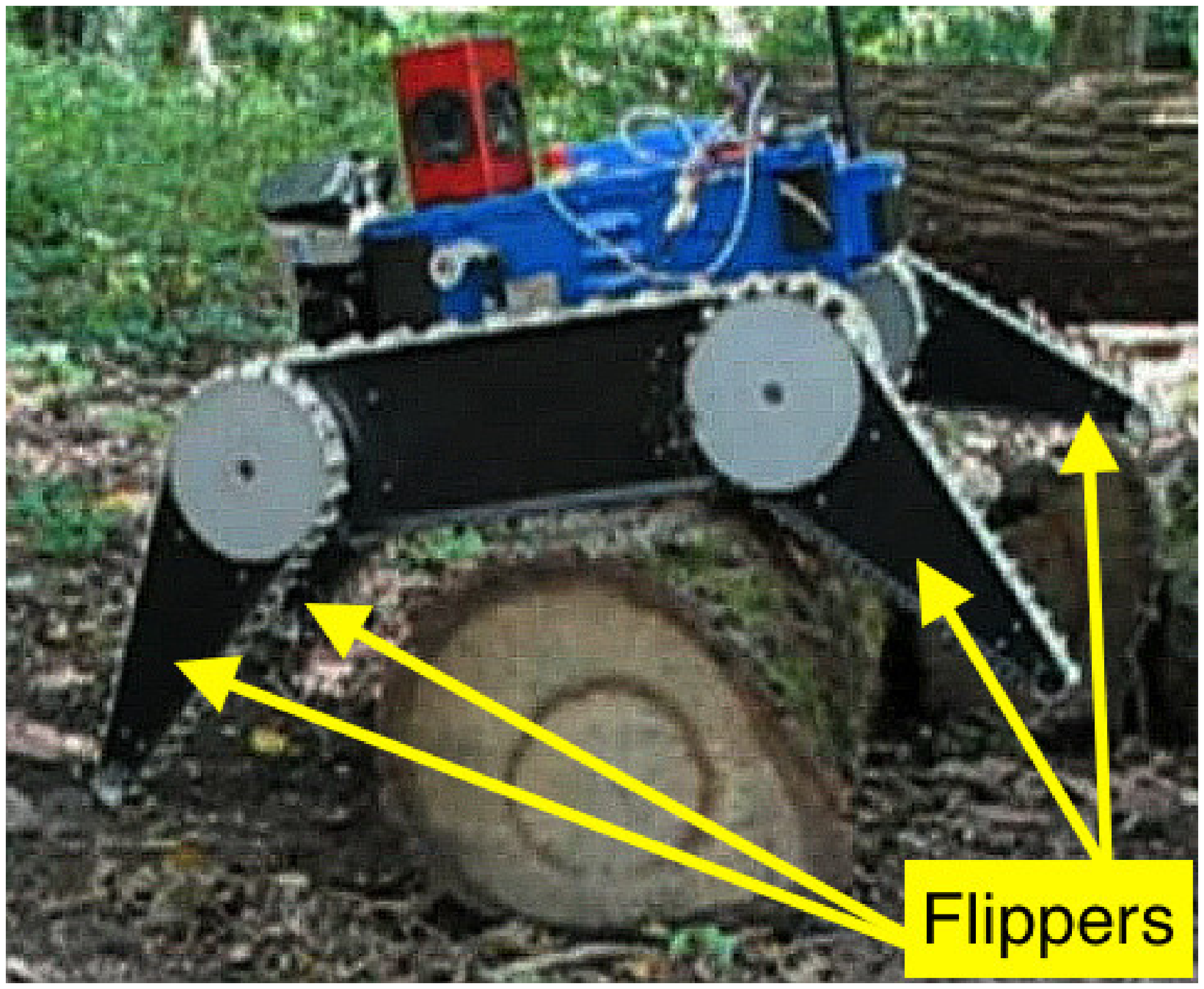}
	\includegraphics[height=0.41\linewidth]{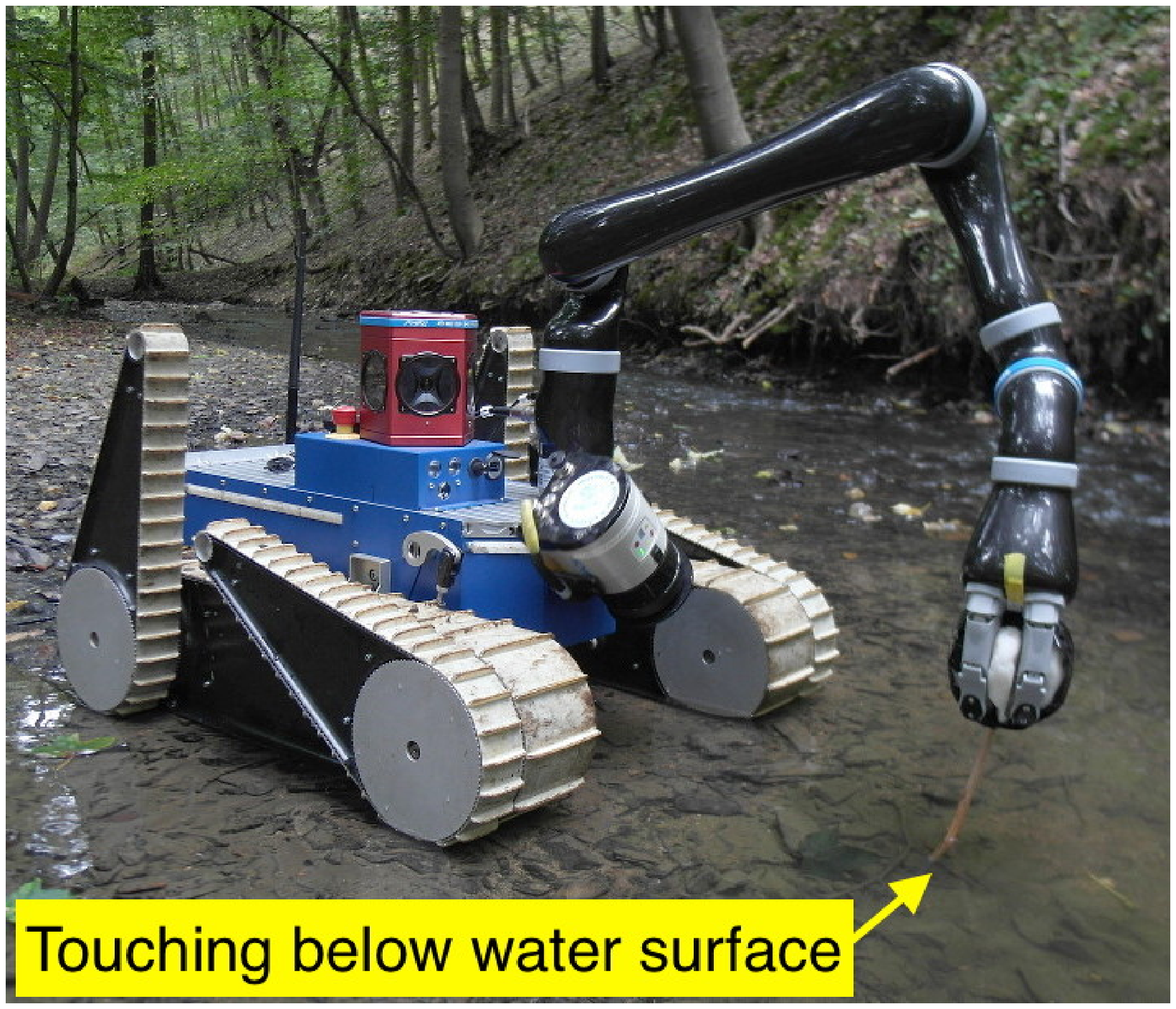}
	\caption{\textbf{Left:} Controlling robot morphology (flippers) allows for traversing obstacles. \textbf{Right:} Robotic arm inspects terrain below water surface compensating thus incomplete lidar measurement.}
	\label{fig:intro}
\end{figure}

\IEEEPARstart{S}{ince} exploration of unknown disaster areas during \emph{Urban Search~\&~Rescue} missions (USAR) is often dangerous, teleoperated robotic platforms are usually used as a suitable replacement for human rescuers.
Motivation to our research comes from field experiments with a tracked mobile robot with four articulated subtracks (flippers, see \autoref{fig:intro}).
The robot morphology allows to traverse complex terrain. A high number of articulated parts brings, however, more degrees of freedom to be controlled. 
Manual control of all available degrees of freedom leads to undesired cognitive load of the operator, whose attention should be rather focused on reaching the higher-level USAR goals. 
To reduce the cognitive load of the operator, the autonomy of the platform has to be increased; however, it still has to fall within the bounds accepted by the operators---a~compromise known as \emph{accepted autonomy} has to be reached\if@shownames~\cite{NIFTi-JAR2014}\fi.

In~\cite{zimmermann-icra2014}, a~Reinforcement-Learning--based \emph{autonomous control}~(AC) of robot morphology (configuration of flippers) is proposed.
Its goal is to allow smooth and safe traversal of complex and previously unknown terrain while letting the operator specify the desired speed vector.
The traversing task is called \emph{Adaptive Traversal}~(AT).
Natural and disaster environments (such as forests or collapsed buildings) yield many challenges that include incomplete or incorrect data due to reflective surfaces such as water, occluded view, presence of smoke, and deformable terrain such as deep snow or piles of rubble.
Since simple interpolation of the missing terrain profile has proved to be insufficient,~\if@shownames we\else Zimmermann~et~al.\fi~presented an improved AC algorithm that better handles incomplete sensory data (using marginalization)~\cite{zimmermann-icra2015}.

In this work, we extend and improve the AC pipeline introduced in \if@shownames~our previously published work\fi~\cite{zimmermann-icra2014,zimmermann-icra2015} (see~\autoref{fig:intro-schema} for an overview). \textbf{The  novel contributions include:}
\textbf{(i)} introducing a~safety measure which allows to invoke tactile exploration of non-visible terrain if needed; 
\textbf{(ii)} several strategies for the tactile exploration with a~body-mounted robotic arm;
\textbf{(iii)} two $Q$-function representations which allow easier marginalization and achieve comparable (or better) results;
\textbf{(iv)} and finally, an extensive experimental evaluation of the Autonomous Control. The real-world experiments cover more than 115 minutes of robot time during which the robot traveled 775 meters over rough terrain obstacles. 

\section{Related work}\label{sec:SOTA}

Many approaches focus on optimal robot motion control in environments with a known map, leading rather to the research field of trajectory planning~\cite{Colas-IROS-2013,Brunner2012,Tsai2011}.
Contrary to planning, AC is useful in previously unknown environments and hence can provide crucial support to the actual procedure of map creation. We rather perceive AC as an independent complement to trajectory planning and not as its substitution.

Many authors~\cite{Gianni-JFR-2014,Colas-IROS-2013,Martin-ISER-2012} estimate terrain traversability only from exteroceptive measurements (e.g. laser scans) and plan the (flipper) motion in advance. In our experience, when the robot is teleoperated, it is often impossible to plan the flipper trajectory in advance from the exteroceptive measurements only.
The reasons are three-fold: (i)~it is not known in advance, which way is the operator going to lead the robot, (ii)~the environment is usually only partially observable, (iii)~analytic modeling of Robot--Terrain Interaction (RTI) in a~real environment is very challenging because the robot can slip or the terrain may deform.
Ho~et~al.~\cite{Ho-IROS-2013} directly predict the terrain deformation only from exteroceptive measurements to estimate traversability. They do not provide any alternative solution when exteroceptive measurements are missing. Abbeel~et~al.~\cite{Abbeel-NIPS-2007} use a different approach---they use only proprioceptive measurements for helicopter control, which often works well for aerial vehicles (unless obstacle avoidance is required). We propose that reactive control based on all available measurements is needed for ground vehicles (where obstacle avoidance or robot--ground interaction is essential).

An ample amount of work~\cite{Weiss2006, Kim2010, DuPont2008} has been devoted to the recognition of traversal-related manually defined classes (e.g. surface type, expected power consumption or slippage coefficient). However, such classes are often weakly connected to the way the robot can actually interact with the terrain.
Few papers describe the estimation of RTI directly. For example, Kim~et~al.~\cite{Kim-Sun2006} estimate whether the terrain is traversable or not, and Ojeda~et~al.~\cite{Ojeda2006} estimate power consumption on different terrain types. In literature, the RTI properties are usually specified explicitly~\cite{Ojeda2006,Ho-ICRA-2013, Kim-Sun2006} or implicitly (e.g. state estimation correction coefficient~\cite{Reinstein2013b, Reinstein-TRO-2013}). 

Since RTI properties do not directly determine the optimal reactive control, their estimation can be completely avoided.
Zhong~et~al.~\cite{hexapod} present a trajectory tracking approach, in which they control a~hexapodal robot and utilize force sensors in the legs to detect unexpected obstacles and walk over them.
The algorithm tries to minimize the trajectory error caused by obstacles, so that the underlying controller does not need to take them into account.
\if@shownames~We \else Zimmerman~et~al. \fi proposed a~different algorithm~\cite{zimmermann-icra2014} that explicitly takes the terrain into account (which should yield better results than trying to hide the terrain from the controller).
The algorithm is based on Reinforcement Learning, which has been successfully used e.g. in learning propeller control for acrobatic tricks with an RC helicopter~\cite{Abbeel-NIPS-2007,Abbeel-ICML-2005}.
Since it is possible to model the helicopter-air interactions quite plausibly, an RTI model can be used to speed up the learning.
In case of ground vehicles, analytical modeling of RTI is very difficult.
Therefore, we rather focus on a model-free RL technique called $Q$-learning (used e.g. to find optimal control in~\cite{Yu2015}). 
In $Q$-learning, state is mapped to optimal actions by taking ``argmax'' of the so-called $Q$~function (the sum of discounted rewards).
In our case, the state space has high dimension (some dimensions with continuous domain), and therefore the $Q$~function cannot be trained for all state--action pairs. Thus, it is modeled either by Regression Forests~(RF) or by Gaussian Processes~(GP). Regression Forests are known to provide good performance when a~huge training set is available~\cite{Shotton-cvpr2011}, with learning complexity linear in the number of training samples. Gaussian Processes present an efficient solution in the context of Reinforcement Learning for control \cite{Deisenrothtobepublished}.

To deal with incomplete data, the $Q$~function values have to be marginalized over missing features. Such marginalization is often tackled by sampling~\cite{Lizotte-NIPS-2008,Tanner-JASA-1987} or EM~algorithm~\cite{Ghahramani-NIPS-1994}. Especially for GPs with Squared Exponential kernel, the Moment Matching marginalization method was proposed by Deisenroth~et~al.~\cite{Deisenrothtobepublished}. Marginalization by Gibbs sampling was evaluated for GPs and piecewise constant functions in~\cite{zimmermann-icra2015}.

We are not aware of~any real mobile platform which would use a~robot arm as an active sensor for inspecting unknown terrain. Most of the efforts in active inference are directed towards active classification~\cite{Doumanoglou-ECCV-2014,Bjorkman-IROS-2013,Jia-BMVC-2010} or active 3D~reconstruction.
Doumanoglou~et~al.~\cite{Doumanoglou-ECCV-2014} use two robotic arms for folding an unknown piece of cloth whose type is recognized from RGBD data (Kinect). One view is usually insufficient, therefore the cloth needs to be turned around to generate an~alternative view. The turning action is implicitly learned with Decision Forests. Bjorkman~et~al.~\cite{Bjorkman-IROS-2013} also recognize objects from RGB-D data. In contrast to~\cite{Doumanoglou-ECCV-2014}, Bjorkman~et~al. use the robotic arm as an active sensor, to touch the self-occluded part of the object in order to reconstruct the invisible 3D shape. 
While all these classification approaches actively evaluate features in order to discriminate the true (\emph{single}) object class from other possible classes as fast as possible, the $Q$-learning--based inference presented here evaluates the features in order to find some of the (\emph{multiple}) suitable flipper configurations that allow for a~safe and efficient traversal.

\section{Overview}\label{sec:overview}

\textbf{$\mathbf{Q}$-learning}: 
The proposed AT solution is adapted from the~RL technique called $Q$-learning (described first to emphasize the differences). 
The first step in the learning process is driving manually the robot over obstacles to collect a~dataset.
The~state $\mathbf{x}$ (e.g. body pitch angle or terrain shape; see~\autoref{sec:AT}) is sampled at regular time intervals $t = 0, 1, \ldots, T$.
At each time instant~$t$, the operator chooses an action $c^t$ (e.g. the desired flipper positions) that allows to go over the obstacle.
After the dataset is collected, each state-action pair $(c^t, \mathbf{x}^t)$ is assigned a~reward $r^t$ reflecting suitability of choosing the action in the given state.

Then the iterative $Q$-learning process starts, which estimates the $q^t$-values that represent the \emph{sum of discounted rewards} the robot can gather by starting in state $\mathbf{x^t}$, executing action $c^t$, and always taking the action leading to maximum~$q$ from the following state onwards~\cite{Watkins1992a}.
The $q^t$ and $Q$ values are computed using the recurrent $Q$-learning formulas~\cite{Kaelbling1996}:
\begin{eqnarray}\label{eq:q_update1}
q_i^t := q_{i-1}^t + \alpha \Big[r^t + \gamma \max\limits_{c'} Q_{i-1}(c',\mathbf{x^{t+1}}) - Q_{i-1}(c^t, \mathbf{x^t}) \Big]
\end{eqnarray}
\begin{eqnarray}\label{eq:Q_computation}
Q_i(c, \mathbf{x}) := \mathrm{mean}(q_i^t\; | \; c^t=c \land \mathbf{x^t}=\mathbf{x}) := \mathrm{mean}(q_i(c, \mathbf{x}))
\end{eqnarray}
where $q_1^t := r^t$, $\alpha\in[0,1]$ is the learning rate and $\gamma\in[0,1]$ is the discount factor.
From the computation above, it follows that $Q_i(c, \mathbf{x})$ is an unbiased estimator of~$E[q_i(c, \mathbf{x})]$.

When the $Q$-learning is done, we denote $Q = Q_i$ and $q^t = q_i^t$, and the optimal action can be computed as:
\begin{eqnarray}\label{eq:best_flipper_pose}
c^\ast(\mathbf{x}) & = &\argmax_c Q(c,\mathbf{x})
\end{eqnarray}

\textbf{QPDF:}
In this paper, we generalize the standard $Q$-learning to an algorithm that learns a~distribution called QPDF instead of the $Q$~function.
For the QPDF (denoted as $p(q|c,\mathbf{x})$) it holds that 
$$Q(c,\mathbf{x}) = E[q(c, \mathbf{x})] = \int q\cdot p(q| c, \mathbf{x})\; dq $$
There are two reasons for modeling the full QPDF: (i)~measuring the safety of flipper configurations and (ii)~marginalization when only incomplete measurements of~$\mathbf{x}$ are available.
In~\autoref{sec:Q-models}, two QPDF models are presented: (i)~Regression Forests and (ii)~Uncertain Gaussian Processes.

Given the QPDF and full feature vector $\mathbf{x}$, the optimal action $c^\ast(\mathbf{x})$ is:
\begin{eqnarray}\label{eq:best_flipper_pose2}
c^\ast(\mathbf{x}) & = &\argmax_c Q(c,\mathbf{x}) = \argmax_c E[q(c, \mathbf{x})] = \nonumber \\
    & = &\argmax_c \int q\cdot p(q| c, \mathbf{x})\; dq
\end{eqnarray}

\begin{figure}[t!]
	\centering
	\includegraphics[width=\linewidth]{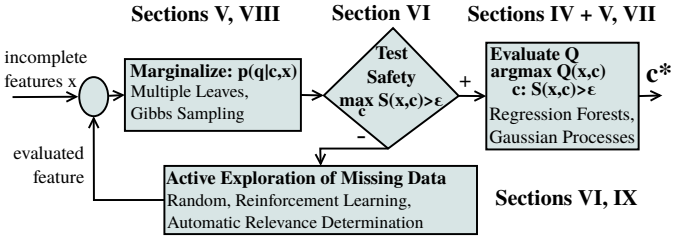}\\ 
	\caption{\textbf{Principle overview:} individual blocks in this scheme correspond to Sections IV-VI.}
	\label{fig:intro-schema}
\end{figure}

\textbf{Missing Data:}
While proprioceptive data are usually fully available, the exteroceptive data are often incomplete. This occurs in case of reflective surfaces such as water or in presence of~smoke. We denote the missing parts of measurements as $\overline{\mathbf{x}}$, and the available measurements as $\mathbf{\widetilde{x}}$, i.e. $\mathbf{x} = [\mathbf{\bar{x}},\mathbf{\widetilde{x}}]$. In the case that $\overline{\mathbf{x}}$ is not empty, $p(q | c,\mathbf{x})$ is marginalized over the missing data $\overline{\mathbf{x}}$ to estimate $p(q | c, \mathbf{\widetilde{x}})$. The marginalization processes for different QPDF models are described in \autoref{sec:marginalization}. Given the marginalized distribution $p(q| c, \mathbf{\widetilde{x}})$ and measurement~$\mathbf{\widetilde{x}}$, the optimal action $c^*$ is estimated by a~small modification of~\autoref{eq:best_flipper_pose2}:
\begin{eqnarray}\label{eq:opt_policy} 
	c^\ast(\mathbf{\widetilde{x}})
	& = &\argmax_c \int q\cdot p(q| c, \mathbf{\widetilde{x}})\; dq.\label{eq:opt-Q}
\end{eqnarray}

Any state-action pair yielding a~negative $q$-value is interpreted as unsafe considering our definition of the reward function\footnote{This assumes the user-denoted penalty for dangerous states to be sufficiently high and discount factor sufficiently different from one; see \autoref{sec:AT} for definition of the reward function.}.
Therefore, the probability that the $q$-value is positive (safe) can be computed, and only sufficiently safe state-action pairs are to be considered further.
The general trend is that the more features are missing, the higher is the scatter of $q$-values. 
Hence, we define the~\emph{safety measure}

\begin{equation}
S(c,\mathbf{\widetilde{x}}) = \int\limits_0^\infty p(q\;|\; c, \mathbf{\widetilde{x}})\;dq,\label{eq:opt-Q-incomplete}
\end{equation}
that corresponds to the probability of achieving a safe state ($q\geq 0$) with action~$c$. Search for the optimal action~$c^*$ (\autoref{eq:opt-Q}) is restricted only to safe actions:
\begin{equation}\label{eq:safety}
S(c,\mathbf{\widetilde{x}})>\epsilon.
\end{equation}

\textbf{Active Exploration:}
If none of the available actions satisfies the safety condition~(\autoref{eq:safety}), the robotic arm is used to measure some of the missing terrain features; see \autoref{fig:intro-schema} for the pipeline overview. In \autoref{sec:TTE}, we propose several strategies that guide the active exploration of missing features in order to find a~safe action as fast as possible. If all terrain features have already been measured and there is still no action satisfying the safety condition, manual flipper control is requested from the operator.

\begin{figure}[ht]
	\centering
	\def\sz{0.5\columnwidth}

	\includegraphics[height=\sz]{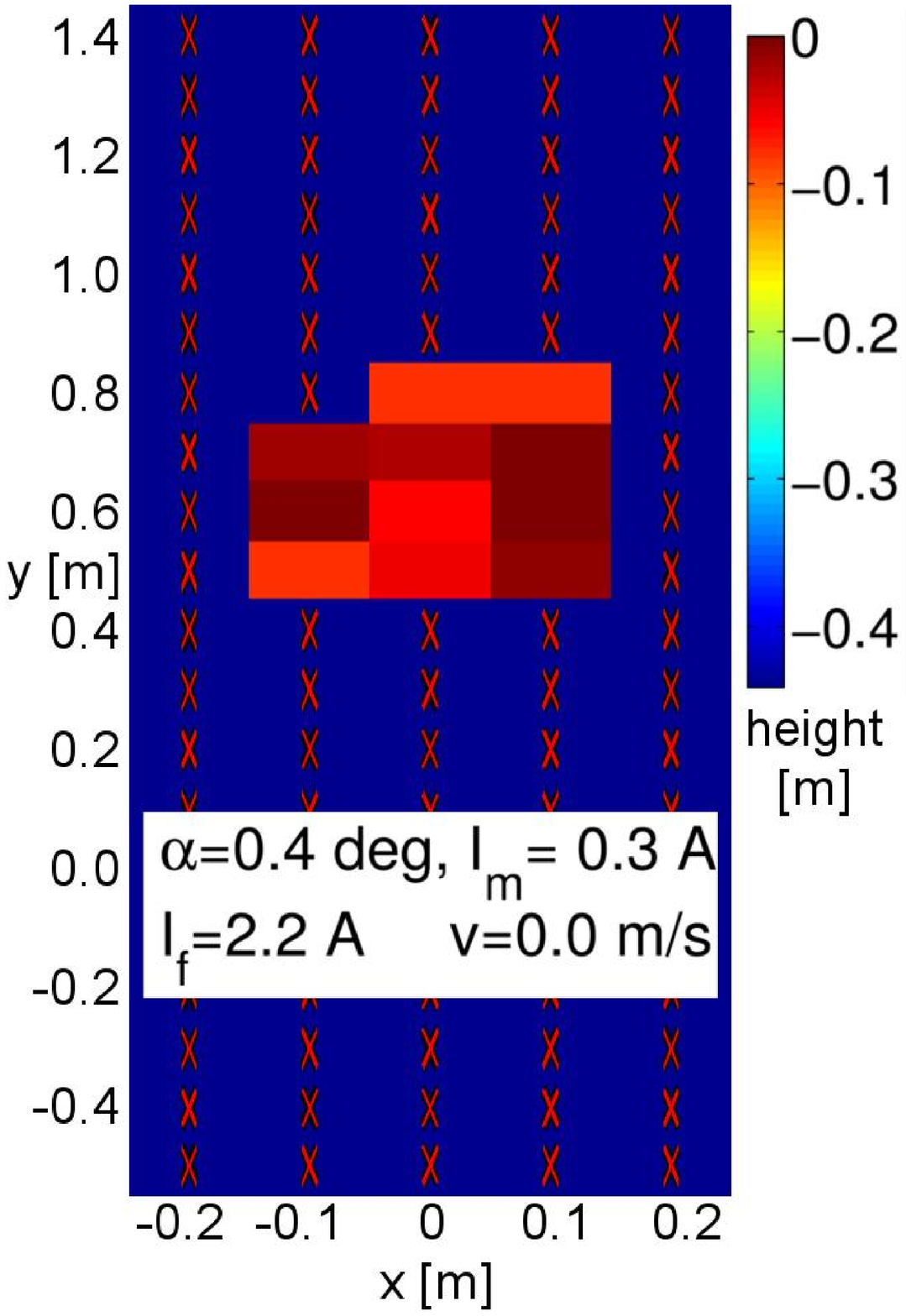}
	\includegraphics[height=\sz]{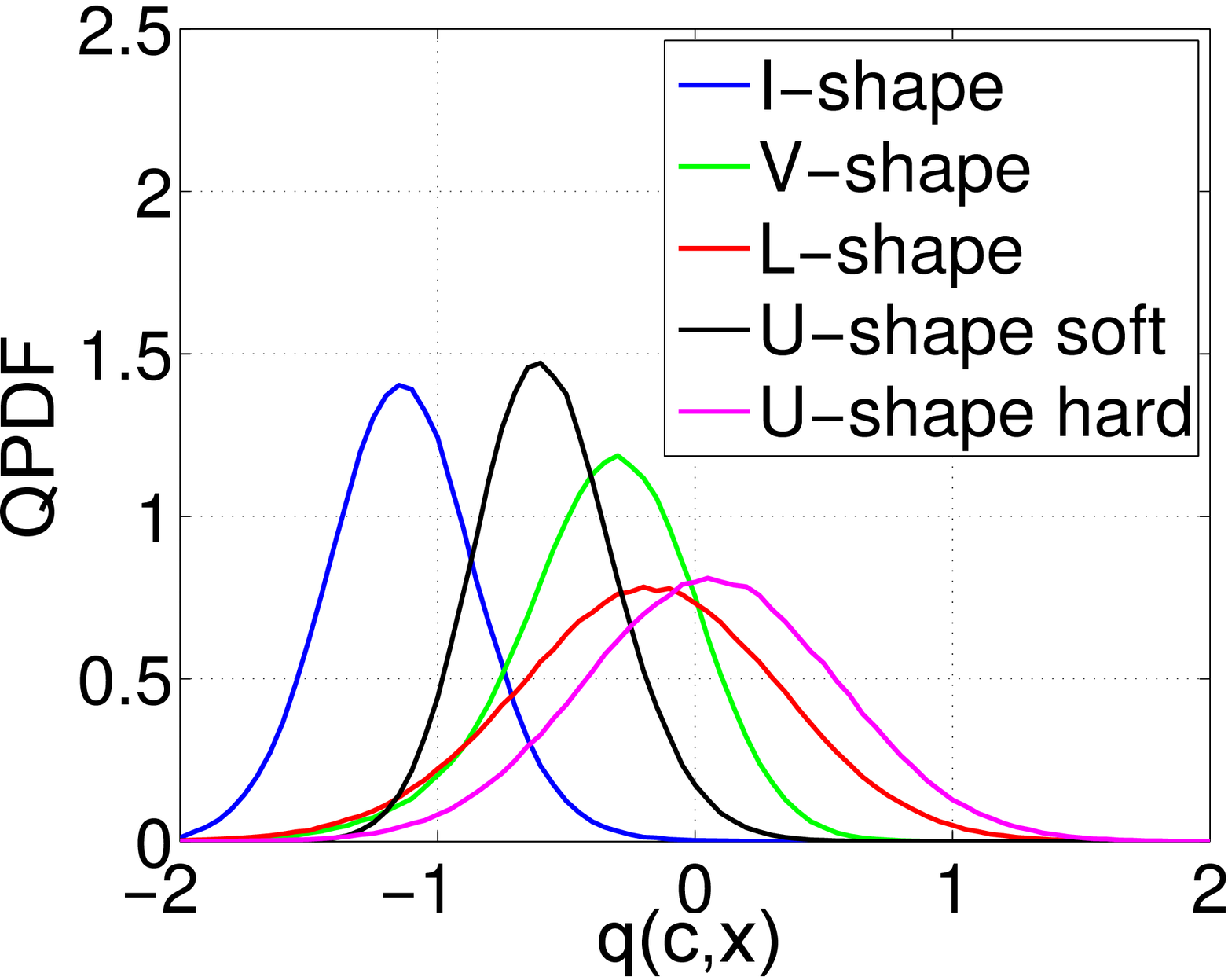}

	\caption{\textbf{ Example of insufficient data. An active exploration is necessary.} The left figure shows the input data; the missing heights in the DEM are outlined by red crosses in a blue rectangle, pitch is denoted by $\alpha$, mean absolute current over both main tracks is denoted $I_m$, mean absolute current in the engines lifting the front flippers is denoted by $I_f$. More details on the features are given in~\autoref{sec:AT}. The right figure contains QPDFs for the five flipper configurations (``*-shape''). The horizontal axis corresponds to the sum of discounted rewards (higher are better), vertical axis contains QPDF. Figure adapted from~\cite{zimmermann-icra2015}.
	}
	\label{fig:exp:qualitative}
\end{figure}

\autoref{fig:exp:qualitative} shows an example situation when active exploration is needed. Looking at the right figure, the highest value of the safety measure $S(c,\mathbf{\widetilde{x}})$ is approximately~0.5. If the safety limit~$\epsilon$ is~$0.8$, tactile exploration is activated, because no action satisfies the safety limit in the current state.

\section{Adaptive Traversability Task}\label{sec:AT}

The AT task is solved for a tracked robot equipped with two main tracks, four independent articulated subtracks (\emph{flippers}) with customizable compliance\footnote{Upper limit of current in the flipper motor used to hold the flipper in position.}, rotating 2D laser scanner (SICK LMS-151), Kinova Jaco robotic arm, and an IMU (Xsens Mti-G); see \autoref{fig:intro}. The task is detailed in the following paragraphs, and a short summary is given in \autoref{tab:model_description}.

\begin{table}[t]
	\centering
	\caption{\label{tab:model_description} Description of the states, actions and rewards}
	{\renewcommand{\arraystretch}{1.3} 
		\begin{tabularx}{\linewidth}{|l|l|X|}
			\hline State & $\mathbf{x}\in\mathbb{R}^n$ & DEM, speed, roll, pitch, flipper angles, compliance, currents in flippers, actual flipper configuration \\
			\hline Actions & $c\in\mathbf{C}=\{1\dots 5\}$ & 5 pre-set flipper configurations~\cite{zimmermann-icra2014} \\
			\hline Reward & $r(c,\mathbf{x}):\mathbf{C}\times\mathbb{R}^n\rightarrow\mathbb{R}$ & $
				\alpha\times\text{user\,reward}\; s_{c,\mathbf{x}}+
				\beta\times\text{pitch\,penalty}+
				\gamma\times\text{roughness\,penalty}
			$ \\
			\hline 
		\end{tabularx} 
	}
\end{table}

\begin{figure}[t]
  \centering
\includegraphics[width=\columnwidth]{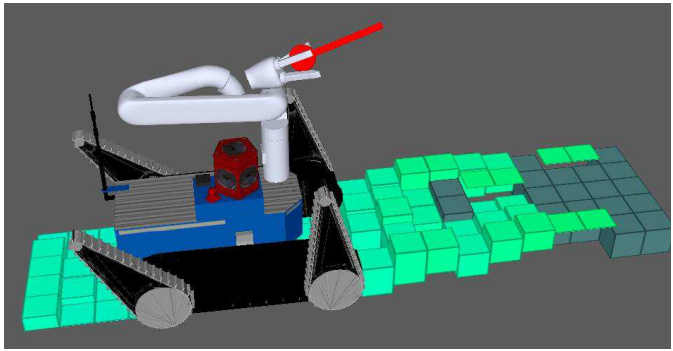}
\caption{\textbf{Digital Elevation Map (DEM):} Example of the DEM representation with dark green used for missing values and light green representing height estimate included in the feature space.}
\label{fig:dem}
\end{figure}

\textbf{States:} The state of the robot and the local neighboring terrain is modeled as
$n$-dimensional feature vector $\mathbf{x}\in\mathbb{R}^n$ consisting of:
\textbf{i)~exteroceptive features:} Individual scans from one sweep of the rotating laser scanner (3~seconds) are put into an Octomap~\cite{hornung13octomap} with cube size of $\mathrm{5\,cm}$. This Octomap is then cropped to close neighborhood of the robot ($\mathrm{50\,cm}\times\mathrm{200\,cm}$ size). Further, the cubes are aggregated into $\mathrm{10\,cm}\times\mathrm{10\,cm}$ columns and mean height in each of these columns is computed. This yields~a local representation of the terrain with $x/y$ sub-sampled to $\mathrm{10\,cm}\times\mathrm{10\,cm}$ tiles (bins) and vertical resolution of $\mathrm{5\,cm}$.
This is what we call a Digital Elevation Map (DEM); see \autoref{fig:dem}.
Heights in the bins are used as exteroceptive features.
\textbf{ii)~proprioceptive features:} Robot speed (actual and desired), roll, pitch, flipper angles, compliance thresholds, actual current in flippers and actual flipper configuration.

\textbf{Actions:} The robot has many degrees of freedom, but only some of them are relevant to the traversal.
The speed and heading of the robot are controlled by the operator. AC is used to control the pose of the four flippers and their compliance, yielding together 8 DOF.
Further simplification of the action space is allowed by observations made during experiments---only 4 discrete (laterally symmetric) flipper configurations are enough for most of the terrain types, and 2 different levels of compliance are also sufficient. The arm has to be in a stable default "transport" position when the robot moves, so its DOFs are ignored.
Finally, 5~\emph{flipper configurations} denoted by $c\in\mathbf{C}=\{1\dots 5\}$ are defined.
These configurations named \emph{I-shape}, \emph{V-shape}, \emph{L-shape}, \emph{U-shape soft} and \emph{U-shape hard} are described in detail in~\cite{zimmermann-icra2014}.

\textbf{Rewards:} The reward function $r(c,\mathbf{x}):(\mathbf{C}\times\mathbb{R}^n)\rightarrow\mathbb{R}$ assigns a real-valued reward for using $c$ in state $\mathbf{x}$. It is expressed as a weighted sum of (i)~user-denoted bipolar penalty $s_{c, \mathbf{x}}$ specifying whether executing $c$ in state $\mathbf{x}$ is \emph{permitted} (safe),
(ii)~high pitch angle penalty (preventing robot's flip-over), and (iii)~the motion roughness penalty measured by accelerometers.

\section{QPDF Representation and Learning}\label{sec:Q-models}\label{sec:marginalization}

In\if@shownames~our previous work\fi~\cite{zimmermann-icra2014} piecewise constant functions were introduced as a~method to represent $Q$~functions.
For the case of missing features, Gaussian Processes with Rational Quadratic kernel were used to represent $Q$~functions in\if@shownames~our following work\fi~\cite{zimmermann-icra2015}.
In the latter work, Regression Forests are trained on fetaures completed by Gibbs sampling marginalization of the missing features.
In this section, we propose two new approaches to QPDF representation that tackle the case of incomplete data.

\subsection{Regression Forests}\label{sec:THEORY:Forest}
The first method is based on Regression Forests with incomplete data on their input, representing the QPDF in their leaves (instead of first estimating the missing features and then computing~$Q$ from a~full feature vector, as the previous method does).
Thus we avoid the unnecessary step of reconstructing the missing features, and can directly use the incomplete input to estimate QPDF.

\textbf{Learning:}
The QPDF for each configuration is modeled independently by a Regression Forest.
The trees are constructed sequentially, always building one until all leaves are \emph{terminal} (see further), and then starting to build another one.
To train each particular tree, a~training set consisting of $m$~training samples $[\mathbf{x}_1,\dots, \mathbf{x}_m]$ is given, with corresponding $q$-values  $[q_1,\dots, q_m]$.
Each training sample $\mathbf{x}_k$ is an $n$-dimensional vector of features $\mathbf{x}_k=[x^1_k\dots x^n_k]^\top$. The tree is built by a~greedy recurrent algorithm, that selects the splitting feature $j^*\in J=\{1\dots n\}$ and split threshold $s^*$.
The splitting feature and threshold are selected to minimize the weighted variance of $q$-values in the left and right sub-tree in each node as follows~\cite{zimmermann-icra2015}:
$$
(s^*, j^*) = \argmin_{(s,j)}\; |R_1(s,j)|\cdot\operatorname*{var\mathit{(q_k)}}_{k\in R_1(s,j)}+ \; |R_2(s,j)|\cdot\operatorname*{var\mathit{(q_k)}}_{k\in R_2(s,j)}
$$
where $R_1(s,j)=\{k\; |\; x_k^j\leq s\}$ is the set of indices descending to the left sub-tree, and $R_2(s,j)=\{k\; |\; x_k^j> s\}$ is the set of indices descending into the right sub-tree. The tree is constructed recursively. If a~stopping criterion is satisfied (either minimum number of samples per node, or tree height), a~\emph{terminal leaf} is created, which contains discretized QPDF histogram (estimated from $q$-values of all training samples that descended to that leaf). Specifically, if the value of the splitting feature is unknown in sample $\mathbf{x}_i$ (e.g. occluded), then it descends into both sub-trees.

\textbf{Marginalization:}
To obtain the marginalized distribution $p(q | c, \mathbf{\widetilde{x}})$, sample $\mathbf{\widetilde{x}}$ is put to the input of the forest.
If a~tested feature is missing in $\mathbf{\widetilde{x}}$, the algorithm descends into both sub-trees similarly to the learning procedure.
The final QPDF is then a~weighted average of histograms in all reached leaves in all trees (properly normalized to be a~distribution).
Weights are given by prior probabilities of leaves estimated from training data.
We call this \emph{Multiple Leaves marginalization}.

\subsection{Gaussian Processes}\label{sec:THEORY:GP}

Gaussian processes~\cite{Deisenrothtobepublished} are the extension of multivariate Gaussians to infinite-size collections of real valued variables and can be understood as joint Gaussian distributions over random functions.
The essential part of GP learning is given by the choice of a kernel function (parametrized by a set of hyper-parameters $\boldsymbol{\theta}$).
We use the common \emph{Squared Exponential} kernel function (SE), for which the \textit{Uncertain} Gaussian Processes are derived in~\cite{Deisenroth-2010}.
This allows processing features with unknown or uncertain values.
In case Uncertain GPs are not necessary, i.e. Gibbs sampling is used to handle uncertain values (as in~\cite{zimmermann-icra2015}), the \emph{Rational Quadratic} (RQ) kernel that performs slightly better than SE can be used. 
Both SE and RQ kernels enable \emph{Automatic Relevance Determination}~\cite{rasmussen2004gaussian}, which can be interpreted as embedded feature selection performed automatically when optimizing over the kernel hyper-parameters $\boldsymbol{\theta}$.
The ARD values are utilized in~\autoref{sec:TTE:ARD}.

\textbf{Learning:}
A~standard regression model is used, assuming the data $\mathcal{D}=\{\mathbf{X}=[\mathbf{x}_1, \dots, \mathbf{x}_m]^T, \mathbf{q}= [q_1, \dots, q_m]^T \}$ were generated according to $q_i=h(\mathbf{x}_i)+\epsilon_i$, where $h:\mathbb{R}^n \rightarrow \mathbb{R}$, and $\epsilon_i \sim \mathcal{N}(0, \sigma_\epsilon^2)$ is independent Gaussian noise.
Thus, there is a direct connection between $h(\mathbf{x})$ and the QPDF.
For each configuration $c$, the Uncertain GP learning procedure is used to train a~GP model that predicts the given $q$-values.
The learning procedure\footnote{Due to the page limitation, the detailed equations are not given here.} is described in detail in~\cite{Deisenroth-2010}.

\textbf{Marginalization:}
GPs consider $h$~as a random function in order to infer posterior distribution $p(h | \mathcal{D})$ over $h$ from the GP prior $p(h)$, the data $\mathcal{D}$, and assumption on smoothness of~$h$~\cite{Deisenroth-2010}.
The posterior is estimated to make predictions at inputs (the testing data) $\mathbf{x} \in \mathbb{R}^n$ about the function values~$h(\mathbf{x})$, which can be used as the QPDF.
Since the posterior is no longer a~Gaussian, it is approximated by a Gaussian distribution, using e.g. the Moment Matching method described in~\cite{Deisenrothtobepublished}.

\section{Tactile Terrain Exploration}\label{sec:TTE}

Given the QPDF, safety condition~(\autoref{eq:safety}) is evaluated for all possible configurations. If more than one safe configuration exists, AC~chooses the one that yields the highest \mbox{$q$-value} mean. If none of the configurations is safe, the robot is stopped and Tactile Terrain Exploration (TTE) is triggered (example situation is depicted in~\autoref{fig:exp:qualitative}). This exploration utilizes the robotic arm to measure the height in DEM bins in which measurements are missing\footnote{The exploration using robotic arm is inherently slow. However, when needed, it is still worth the extra time.}. The arm actively explores the missing heights until the safety condition~(\autoref{eq:safety}) is satisfied for at least one configuration, or there are no more missing heights (we refer to both these cases as \emph{final states}). If the state in the latter case is still unsafe, the operator is asked to control the flippers manually.

We propose several TTE strategies. The simplest TTE strategy selects the bin to be explored randomly from the set of all missing bins---we refer to this strategy as \textbf{Random}. Further, we propose and evaluate also two better TTE strategies: (i)~the Reinforcement-Learning--based strategy trained on synthetically generated training exploration roll-outs (further referred to as \textbf{RL}~strategy), and (ii)~a~strategy based on \emph{Automatic Relevance Determination} coefficients for QPDFs modeled by the GP (further referred to as the \textbf{ARD}~strategy).

\subsection{RL from Synthetically Generated Training Set}\label{sec:TTE:RL}

The Reinforcement-Learning--based TTE learns a~policy that minimizes the number~$n$ of tactile measurements needed to satisfy the safety condition. In our implementation, a~\textbf{state} is the union of the state used in the AT task (i.e. the proprioceptive and exteroceptive measurements), and the binary mask denoting DEM bins with missing heights. \textbf{Actions} are discrete decisions to measure the height in particular bins. \textbf{Rewards} equal zero until a~final state is reached. In the final state, the roll-out ends and a~reward equal to~$1/n$ is assigned (i.e. the longer it takes, the lower the reward).

Since it is not easy to collect sufficient amount of real examples with naturally missing features, we generate training samples from the real data with synthetically occluded DEMs. The active exploration policy is thus trained by revealing the already known (but synthetically occluded) heights. The $Q$-learning algorithm learns the strategy in several episodes. The initial training set is generated by simulating thousands of TTE roll-outs with the \emph{Random} strategy. The $Q$~function is modeled by a~Regression Forest similar to the one used in \autoref{sec:Q-models} (but this $Q$~function is different from the one used for Autonomous Control!). Once the $Q$~function is learned, the corresponding strategy is used to guide training data collection in the following episode by the DAgger algorithm~\cite{Ross-ICML-2012}. In each episode of the DAgger algorithm, the learned policy is used to select bins just with $0.5$~probability, otherwise the Random strategy is used (which supports exploration in the policy space). After each episode, the policy is updated using the $Q$-learning recurrent formula~(\autoref{eq:q_update1}).

\subsection{ARD for Gaussian Processes}\label{sec:TTE:ARD}
In~\autoref{sec:THEORY:GP}, it is mentioned that both SE and RQ kernels allow for \emph{Automatic Relevance Determination} (ARD), which acts as feature selection.
The ARD values are computed during kernel hyper-parameters optimization (when training the GP), so no extra computing power is needed.
When the learning is done, for each dimension (feature) $d$ of the input data, we have a number $ARD(d)$ that describes how much this dimension influences the output of the GP (lower values mean higher importance).
The TTE strategy utilizing ARD values is as follows:
\textbf{i)}~estimate QPDFs using all GP models,
\textbf{ii)}~select the action (GP model) with the highest $Q$-value (QPDF mean),
\textbf{iii)}~in this GP, compare $ARD(d)$ values for all DEM bin features that are missing in the current state, and choose the bin with the minimum $ARD(d)$ value,
\textbf{iv)}~the chosen bin is then explored using the arm.
This corresponds to choosing the missing feature whose value, if known, maximally influences the QPDFs.

\section*{EXPERIMENTS}

Experimental evaluation is divided into three sections. In~\autoref{sec:exp:AT}, we test the ability of AC to decrease cognitive load of human operators while maintaining roughly the same or better performance. Experiments in~\autoref{sec:exp:marg} demonstrate that if the DEM is partially occluded, the proposed  method yields better results than the previous methods. Last, \autoref{sec:exp:TTE} compares Random, ARD and RL methods for tactile exploration.

In the experiments, different $Q$~function/QPDF representations are denoted by \textbf{PWC} for piecewise constant function proposed in~\cite{zimmermann-icra2014}, \textbf{GP-RQ} stands for Gaussian Processes with Rational Quadratic kernel used in~\cite{zimmermann-icra2015}, \textbf{GP-SE} denotes the Uncertain GPs with Squared Exponential kernel, and finally the Regression Forests defined in~\autoref{sec:Q-models} are referred to as \textbf{Forest}.
The PWC and GP-RQ models can be used either with Least Squares (\textbf{LSq}) interpolation of missing features, or with \textbf{Gibbs} sampling used to marginalize the $Q$~function over the missing data. Regression Forests utilize the \textbf{Multiple Leaves marginalization}.

A~metric called \emph{success rate} is used throughout the experiments to measure the traversal performance both on training data (in the learning phase) and on test data.
It requires that the bipolar manually-assigned part of reward $s_{c,\mathbf{x}}$ defined in~\autoref{sec:AT} is assigned for all actions in all states in the dataset (not just for a~single action, as is required for the learning).
The success rate denotes the ratio of states, in which the AC algorithm selects one of the desired (safe) configurations. Formally:
\begin{eqnarray}\label{eq:success_ratio}
\text{success rate}(\mathbf{X}) = \frac{|\{\mathbf{x}\in\text{X}: c=c^\ast (\mathbf{x}) \land s_{c,\mathbf{x}} = 1\}|}{|\mathbf{X}|}
\end{eqnarray}
where $\mathbf{X}$ is a~set of states, and $c^\ast(\mathbf{x})$ is the optimal configuration from~\autoref{eq:best_flipper_pose} or~\autoref{eq:opt_policy} (depends on the used AC algorithm).

\section{Autonomous Control for Teleoperation}\label{sec:exp:AT}

\begin{table*}
\centering

\caption{\label{tab:at_evaluation}Experimental evaluation of Adaptive Traversability}

\def\experimentwidth{0.22\textwidth}
\def\imagetop#1{\raisebox{-0.95\height}{#1}}
	
\subfloat[
	\textbf{Overall statistics of the experiments.}
	The computation of penalties is described in \protect\subref{fig:atEvaluationDetails}.
	High penalties for experienced operator with 3rd person view (TPV) are given by the fact that an experienced operator allows himself to drive harsher to finish the task faster.
	]{
\begin{tabular}{c@{}c@{}c@{}c@{}c}
	& \textbf{Palette} & \textbf{Stairs} & \textbf{Rubble} & \textbf{Forest} \\
		
	\raisebox{-1.18\height}{\includegraphics[height=0.3\textwidth]{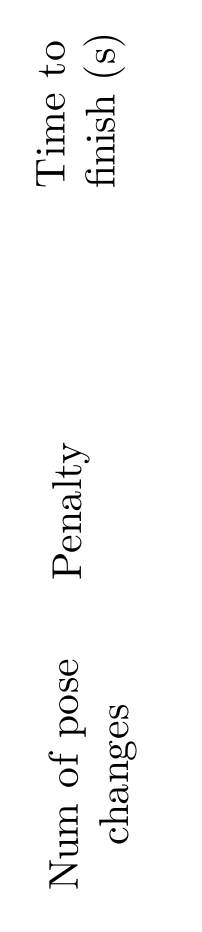}}
	& \imagetop{\includegraphics[width=\experimentwidth]{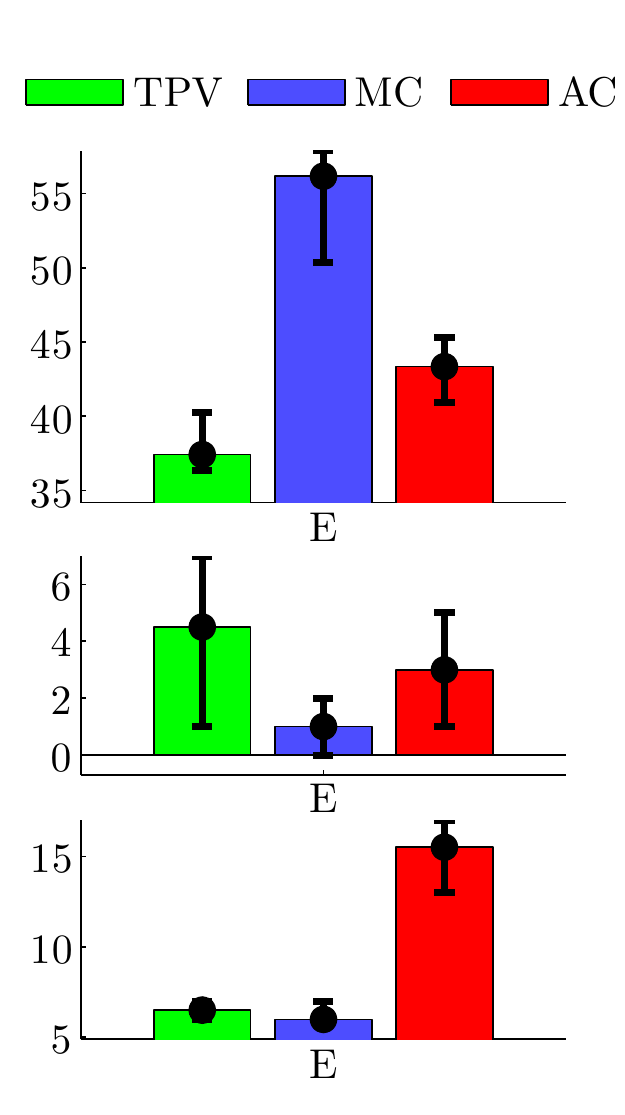}}
	& \imagetop{\includegraphics[width=\experimentwidth]{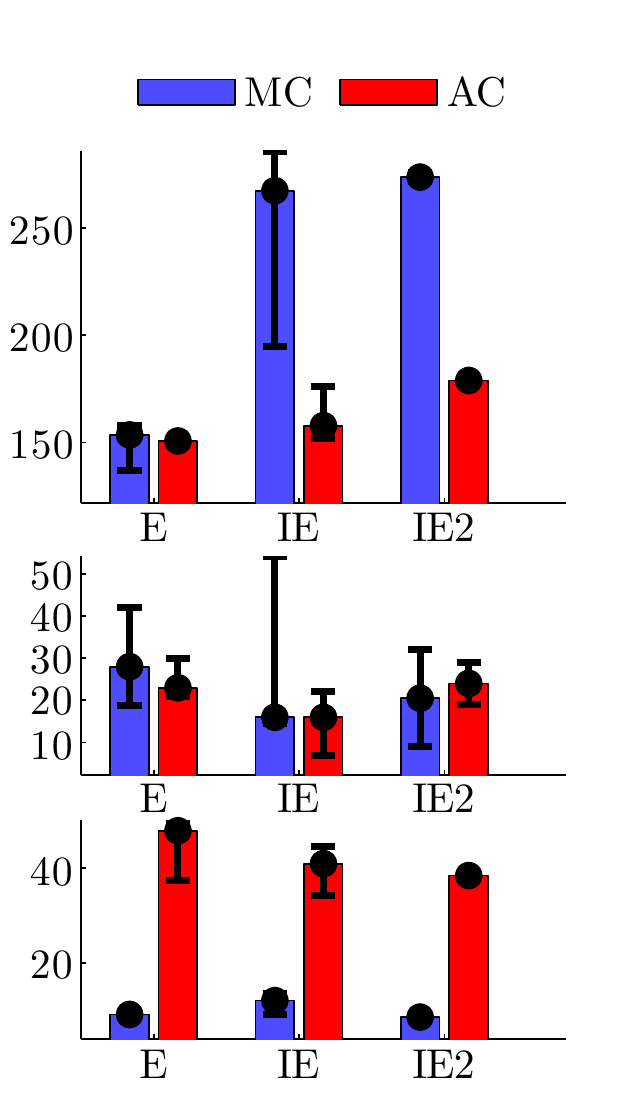}}
	& \imagetop{\includegraphics[width=\experimentwidth]{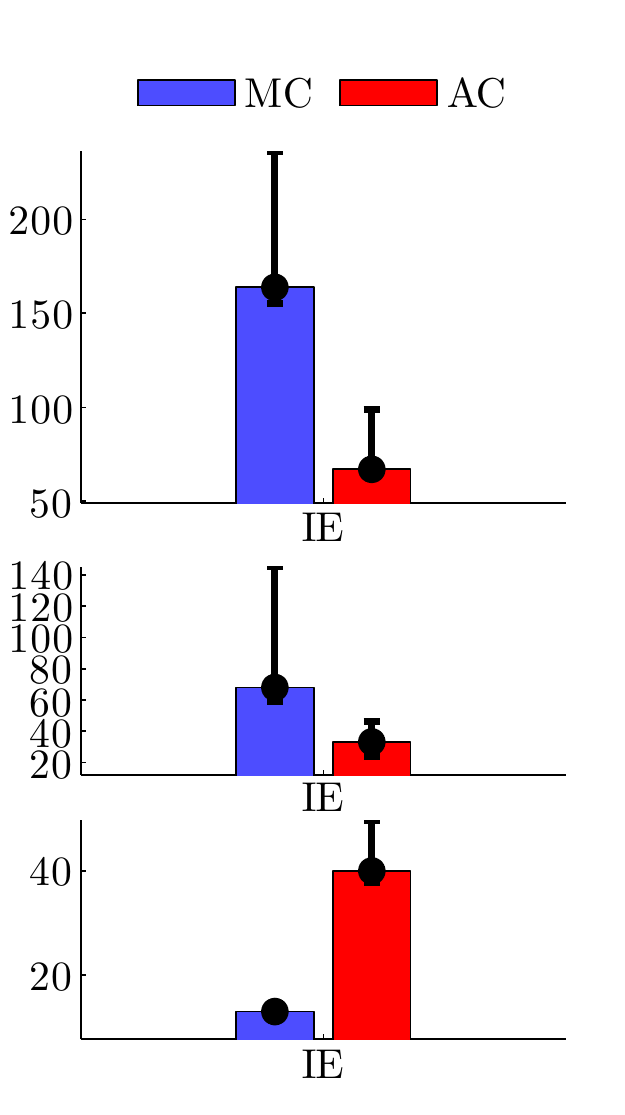}}
	& \imagetop{\includegraphics[width=\experimentwidth]{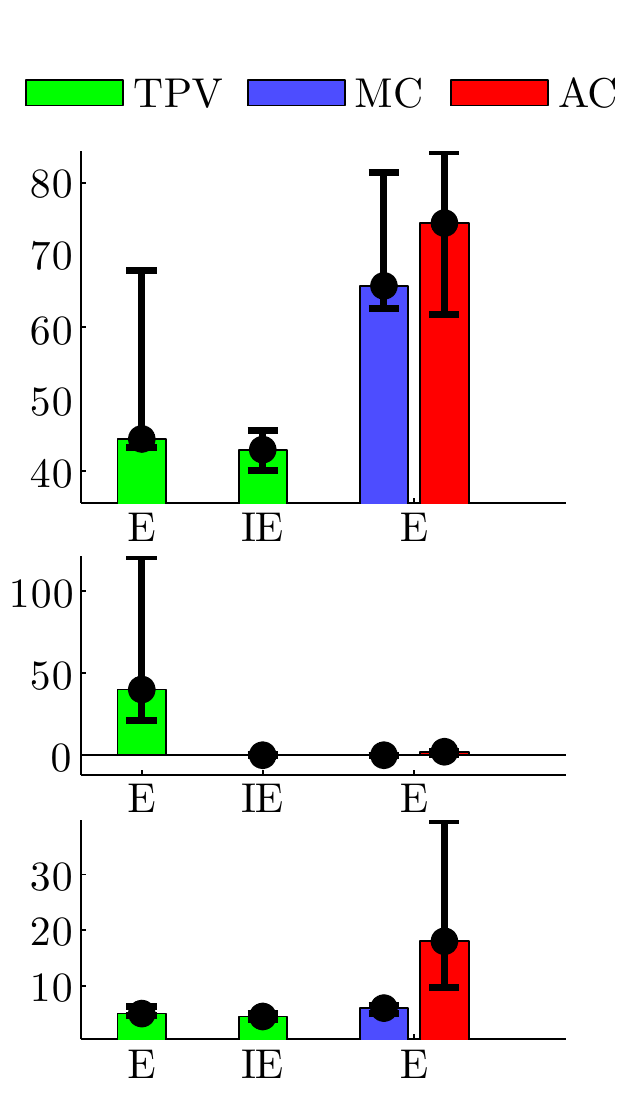}}
	\\
\end{tabular}
	\label{fig:atEvaluationStats}
}

\subfloat[
	\textbf{Penalty details.}
	The horizontal axis always denotes distance traveled during the experiment.
	Dashed lines in \emph{Pitch} and \emph{Acceleration} show the thresholds (\SI{0.5}{\radian} or \SI{2.5}{\meter\per\second\squared}) for counting a \emph{penalty point} (which are plotted in~\autoref{fig:atEvaluationStats}).
	\emph{Acceleration} reflects the ``roughness of motion'' (the higher it is, the worse for the mechanical construction of the robot).
	It is computed as $\sqrt{a^2_x + a^2_z}$ and is averaged over \SI{0.2}{\second} intervals (where $a_x$ is the horizontal acceleration perpendicular to robot motion, and $a_z$ is vertical acceleration with gravity subtracted).	
	]{
\begin{tabular}{c@{}c@{}c@{}c@{}c}	
	& \textbf{Palette} & \textbf{Stairs} & \textbf{Rubble} & \textbf{Forest} \\
	
	\raisebox{-1.0\height}{\includegraphics[height=0.53\textwidth]{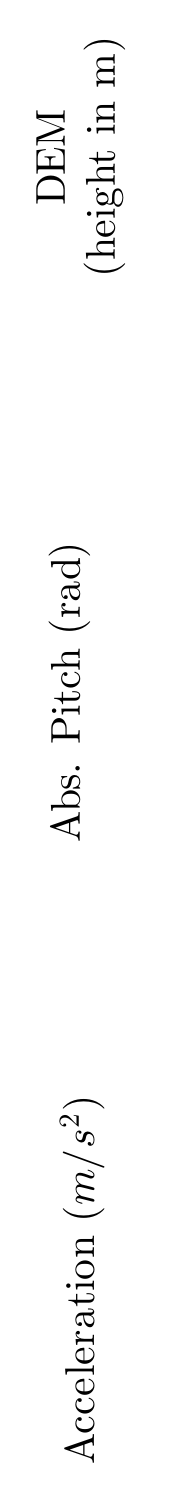}}
	& \imagetop{\includegraphics[width=\experimentwidth]{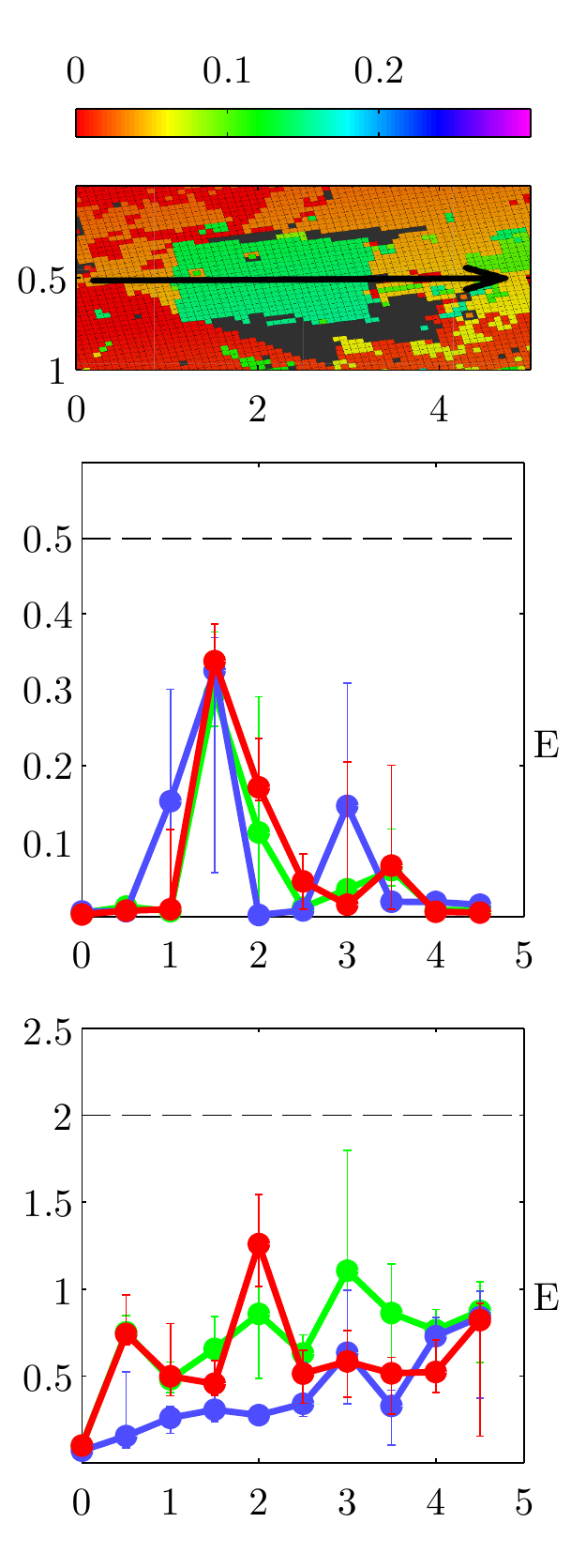}}
	& \imagetop{\includegraphics[width=\experimentwidth]{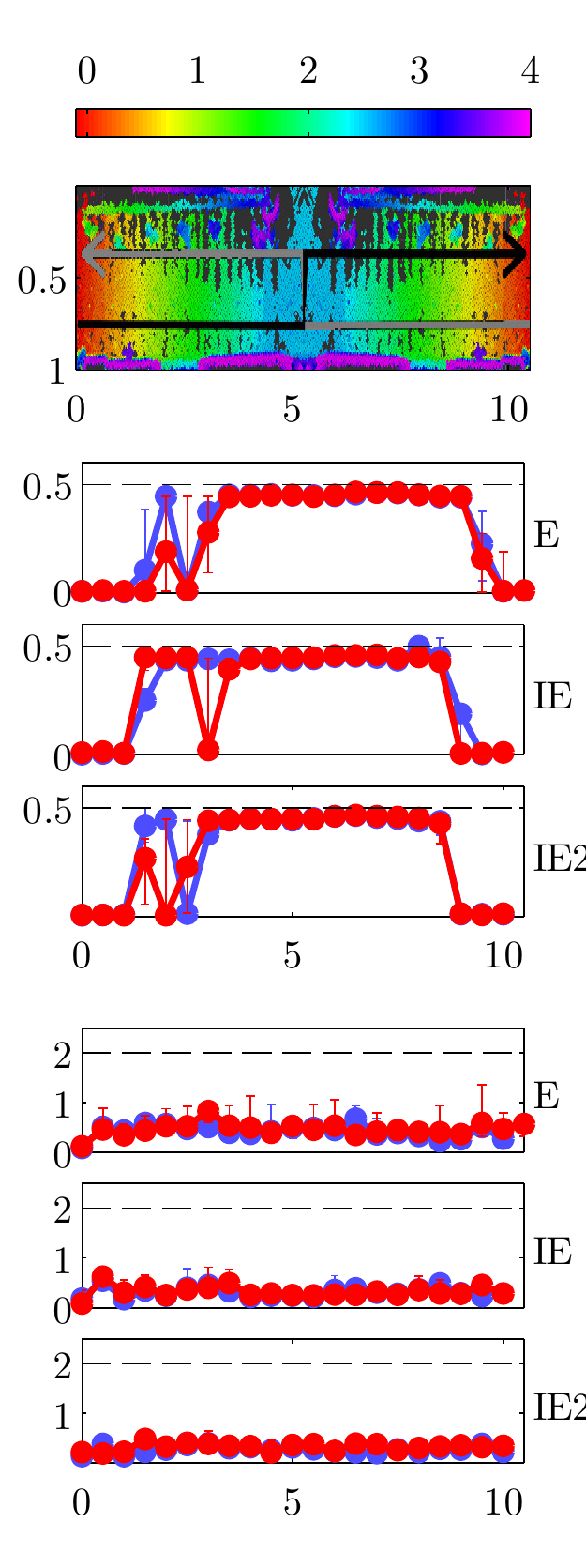}}
	& \imagetop{\includegraphics[width=\experimentwidth]{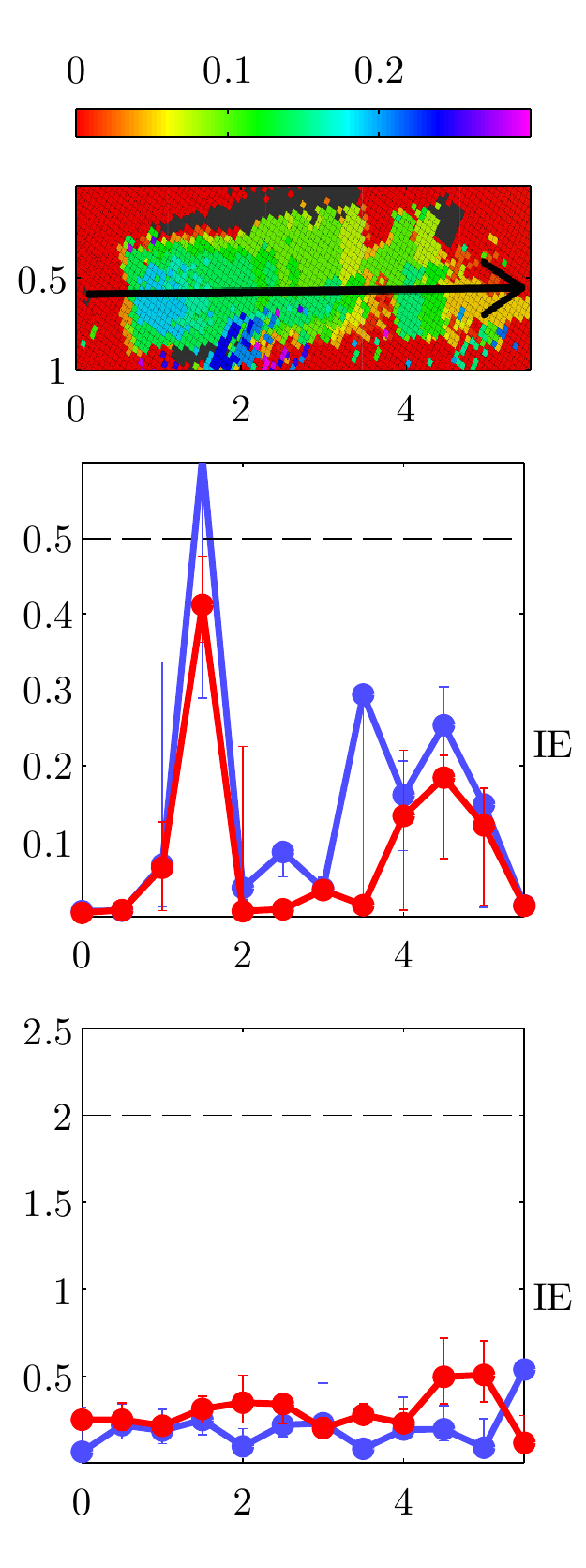}}
	& \imagetop{\includegraphics[width=\experimentwidth]{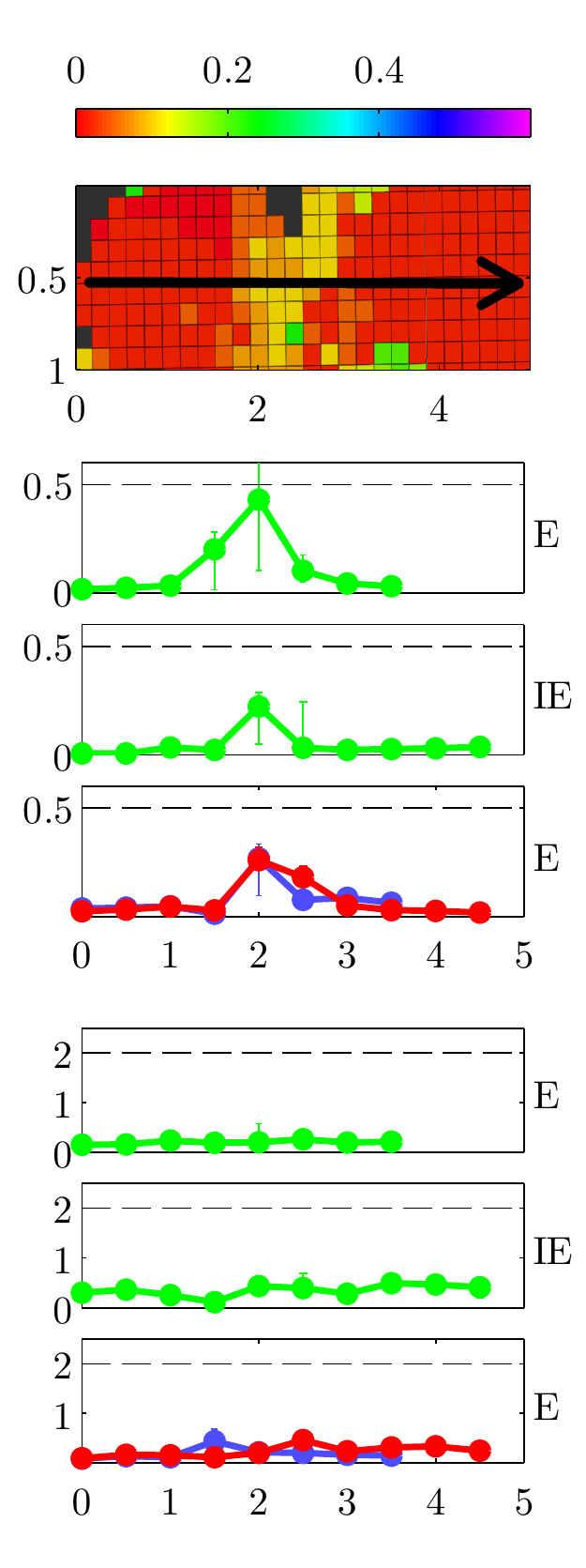}}
	\\
\end{tabular}
		\label{fig:atEvaluationDetails}
}

\end{table*}

\begin{figure}[t]
			\def\imgwidth{0.45\columnwidth}
			\begin{tabular}{l@{\hspace{2pt}}l}
				\raisebox{-\height}{\includegraphics[width=\imgwidth]{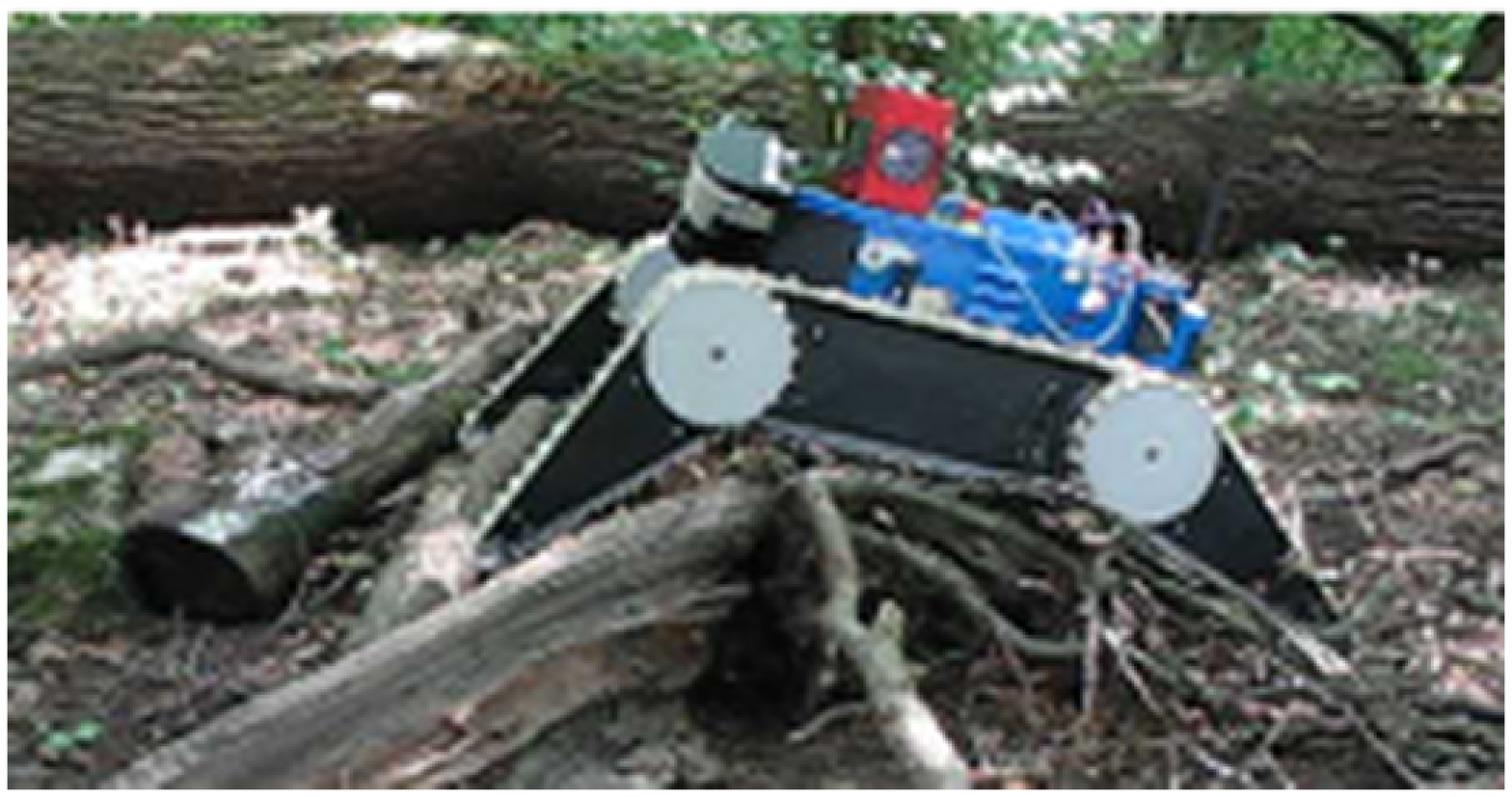}} &
				\raisebox{-\height}{\includegraphics[width=\imgwidth]{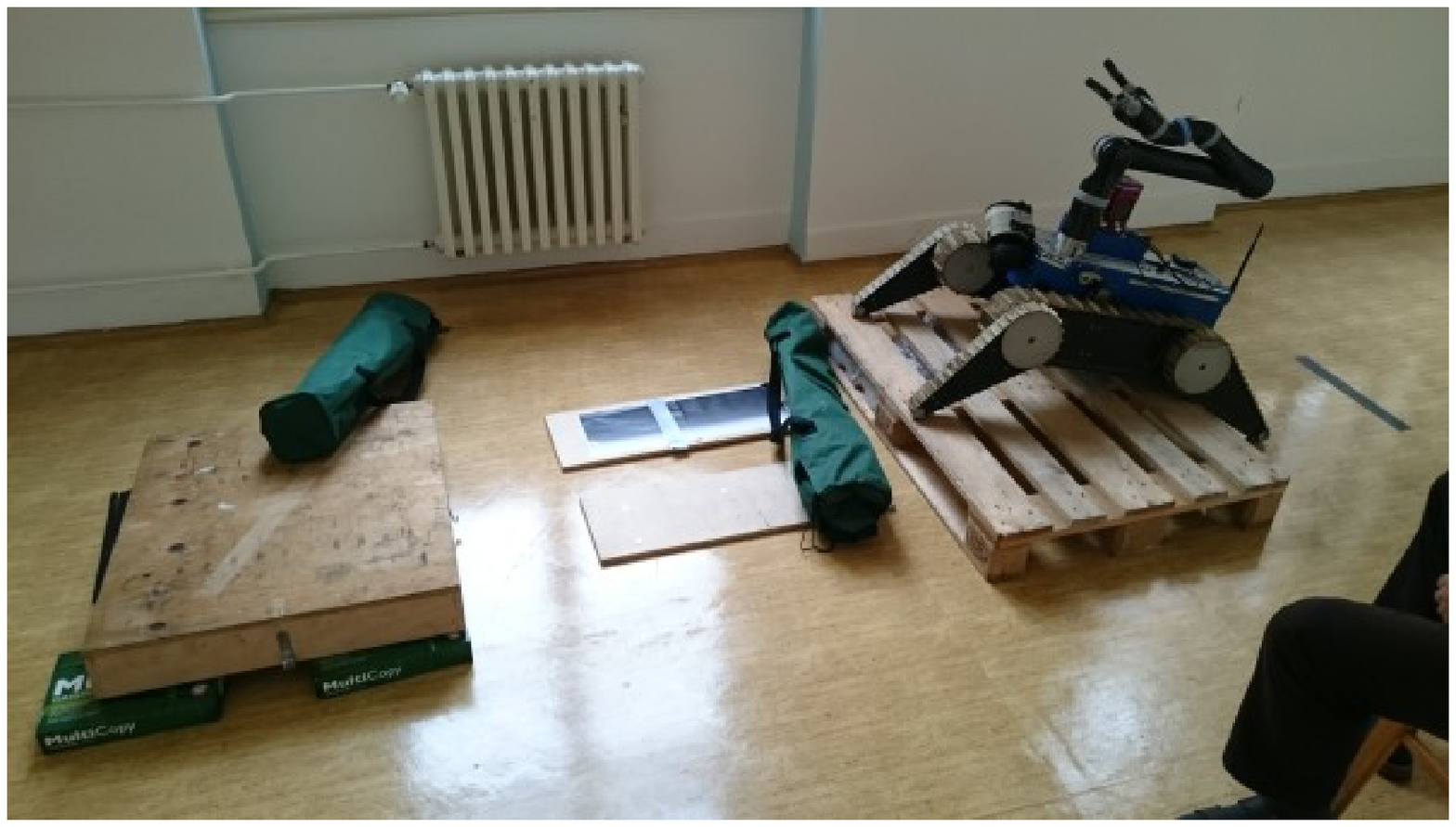}} \vspace{2pt} \\
				
				\includegraphics[width=\imgwidth]{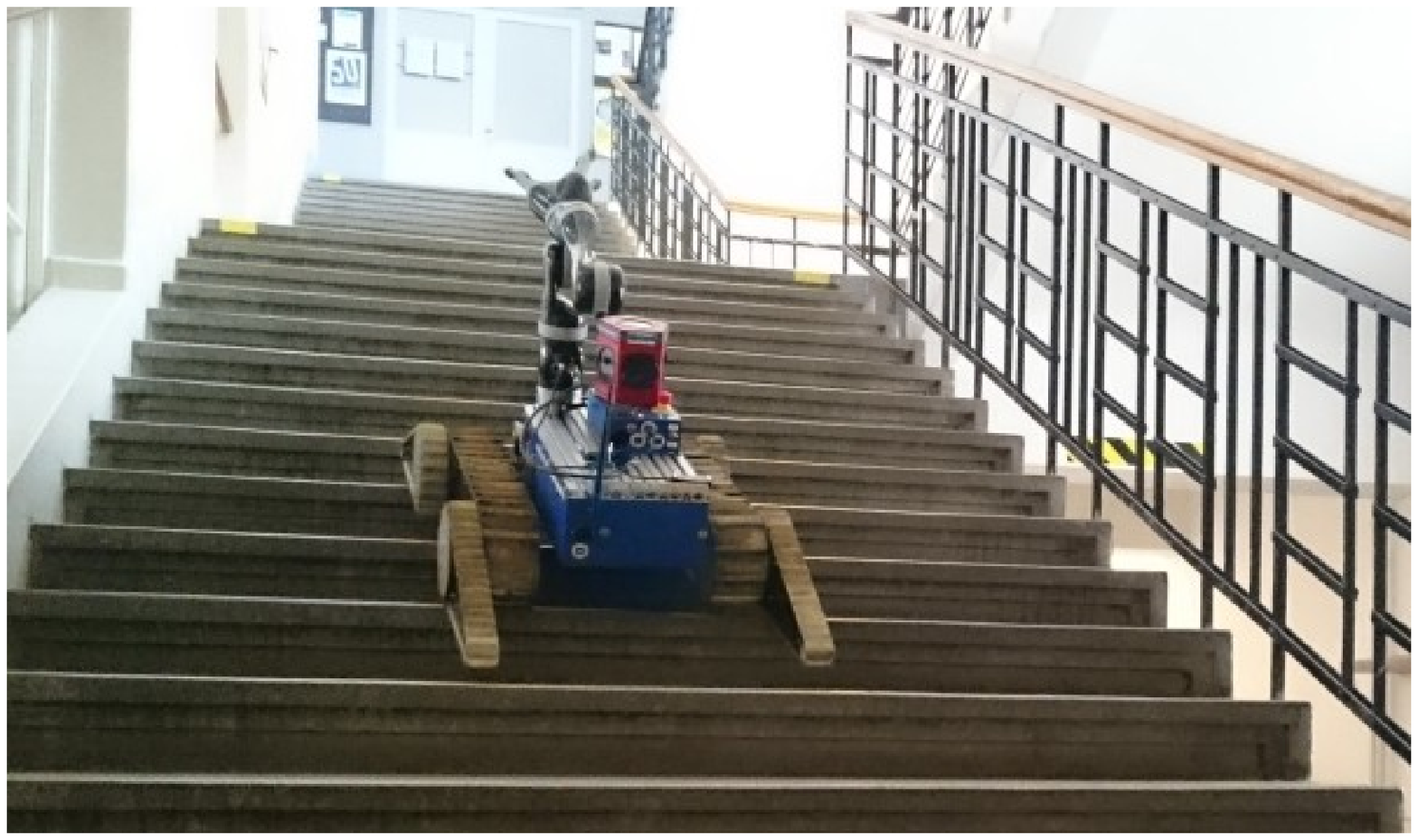} &
				\includegraphics[width=\imgwidth]{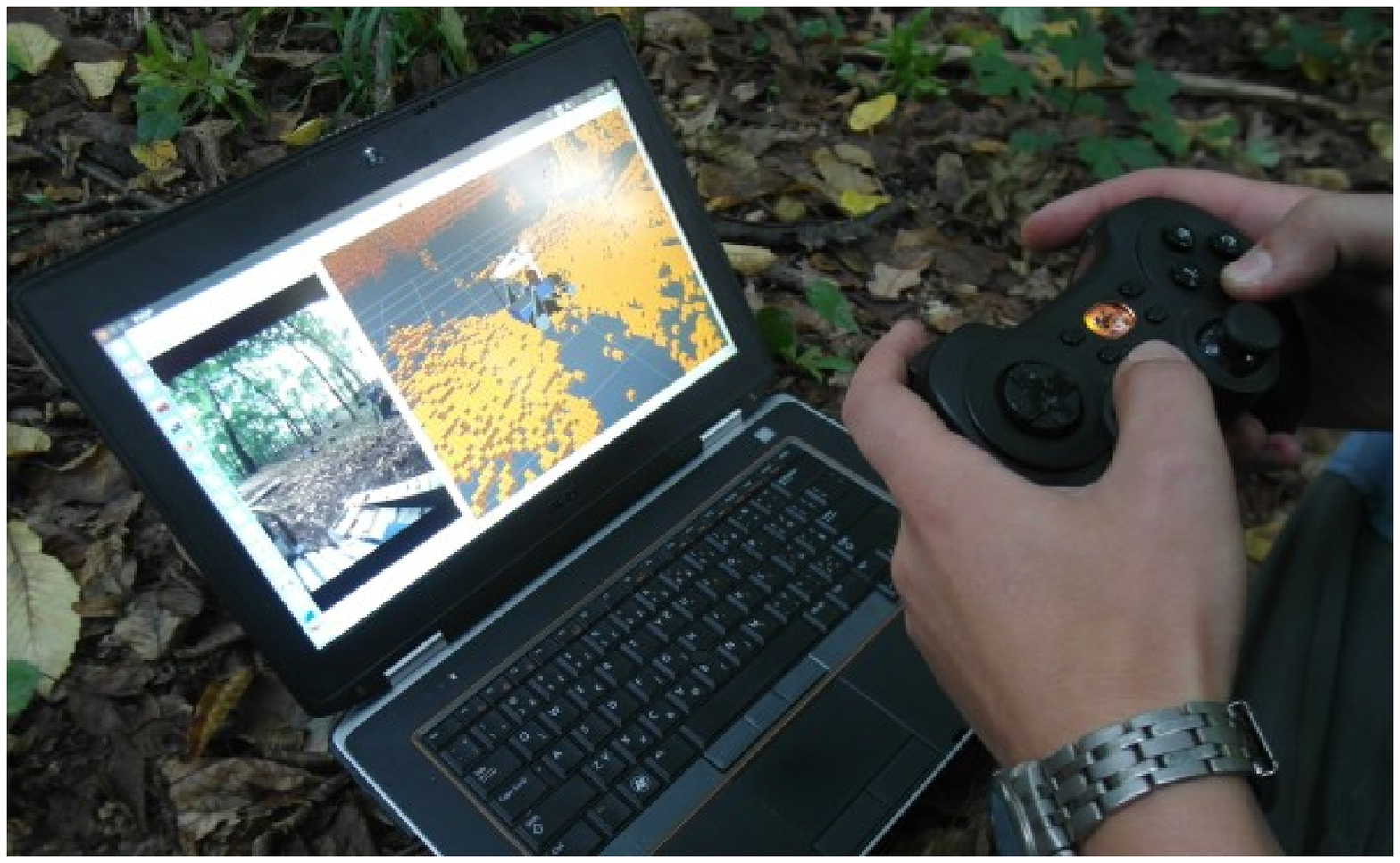}
			\end{tabular}
			\caption{Top left: \emph{Forest} obstacle. Top right: \emph{Rubble} obstacle. Bottom left: \emph{Stairs} obstacle. Bottom right: Operator controlling the robot using only sensor data.
				}
			\label{fig:trn_objects:tst}			
\end{figure}

We evaluate performance of the AC algorithm (without tactile exploration) on a~large dataset comprising of 8~different obstacles (some of them depicted in \autoref{fig:trn_objects:tst}) in 3~types of environment (forest, stairs, hallway) with the robot driven by 3~different operators in both MC and AC modes\footnote{See the attached multimedia showing the test drives.}.
Each of the traversals is repeated 3-10~times to allow for statistical evaluation.
The operators driving the robot are denoted as \textbf{E}~(Experienced), \textbf{IE}~(InExperienced) and \textbf{IE2}~(InExperienced~\#2).
The experiments cover more than 115~minutes of robot time during which the robot traveled over 775~meters.

Experiments in this section only show the results achieved with Regression Forests; other $Q$~function representations were tested in~\cite{zimmermann-icra2014,zimmermann-icra2015}, and Uncertain Gaussian Processes were only tested together with the tactile exploration (see~\autoref{sec:exp:TTE}), since without TTE they performed worse than the Regression Forests (and for creating such a~large dataset, we had to choose one method).

\subsection{Training Procedure}\label{sec:exp_training}

The algorithm was trained in controlled lab conditions using two artificial obstacles created from EUR pallets\footnote{Type EUR~1: 800$\times$1200$\times$\SI{140}{\mm}, see \url{en.wikipedia.org/wiki/EUR-pallet}} and a staircase. The first obstacle is just a single pallet and the second one is a simple simulated staircase composed from three pallets.

We trained the QPDFs represented by RF (one QPDF per flipper configuration) using the algorithm described in \autoref{sec:Q-models}.
Except the standard learning validation metrics, we also evaluated the \emph{success rate} (\autoref{eq:success_ratio}).
We trained the RF QPDF model, and we accomplished a~success rate of \SI{97}{\percent} (which is shown in~\autoref{fig:exp:quantitative}).

\subsection{Testing Procedure}\label{sec:exp_testing}

Each obstacle was traversed multiple times with both manual (\textbf{MC}) and autonomous flipper control (\textbf{AC}) using RFs following \autoref{eq:opt_policy}, and the sensed states contained naturally missing DEM features.
We emphasize that the complexity of testing obstacles was selected in order to challenge robot hardware capabilities.
See the examples in \autoref{fig:trn_objects:tst} and the elevation maps (DEM) of testing obstacles computed online by the robot in \autoref{fig:atEvaluationDetails}.

There is an additional mode called \textbf{TPV} (\emph{Third Person View}) in which the operator had not only the robot sensory data available, but he directly looked at the robot (thus having much more information than the robot can get).
Except for the TPV mode, the operators were only allowed to drive the robot based on data coming from the robot sensors (3D~map + robot pose from sensor fusion~\cite{kubelka2015robust}), which should accomplish a fair comparison of AC and MC.
The TPV mode should be treated as a sort of baseline---it is not expected that AC or MC could be better than TPV in all aspects.

To compare AC and MC quality, three different metrics were proposed and evaluated: 
(i)~traversal time (start and end points are defined spatially),
(ii)~a~sum of pitch angle penalty and roughness of motion penalty, and 
(iii)~the number of flipper configuration changes (which increases cognitive load of the operator in MC, and with the current manual controller, it also takes approx. \SI{1}{\second} to change the flipper configuration and the robot has to be stopped).
\autoref{fig:atEvaluationStats} and \autoref{tab:at_all_results} show quantitative evaluation of some of the experiments. \autoref{fig:atEvaluationStats} depicts 4 out of 8 experiments carried out to verify performance of AC using the best method found---Regression Forests with Multiple Leaves marginalization. All errorbars denote quartiles of the measured values and the circles are in the positions of medians.

\autoref{tab:at_all_results} summarizes all MC/AC experiments (excluding TPV mode experiments, since they should not be compared with MC/AC).

\begin{table}[ht!]
\def\mc#1#2{\multicolumn{2}{l#1}{\textbf{#2}}}
\def\b#1{\textbf{#1}}
\def\arraystretch{1.15}
\caption{\label{tab:at_all_results}Evaluation of all AT experiments}

\setlength{\tabcolsep}{0.3em} 
\begin{tabular}{|l|l||r|r||r|r||r|r|}
\hline \textbf{Obstacle} 			& \textbf{Operator}	& \mc{||}{Time to finish [s]}& \mc{||}{Penalty}		& \mc{|}{Pose changes}	\\
\hline								&					& \b{MC}	& \b{AC}		& \b{MC}	& \b{AC}	& \b{MC}	& \b{AC} 	\\
\hline\hline \textbf{Pallet long} 	& E					& 56.2		& \b{43.3}		& \b{1}		& 3			& \b{6}		& 16		\\
\hline \textbf{Pallet short}		& E					& 41.0		& \b{39.3}		& \b{2}		& 4			& \b{6}		& 15		\\
\hline \textbf{Stairs}				& E					& 154.0		& \b{150.6}		& 28		& \b{23}	& \b{9}		& 48		\\
\hline								& IE				& 267.3		& \b{157.9}		& \b{16}	& \b{16}	& \b{12}	& 41		\\
\hline								& IE2				& 273.7		& \b{178.8}		& \b{21}	& 24		& \b{9}		& 39		\\
\hline \textbf{Rubble 1}			& IE				& 164.0		& \b{66.9}		& 68		& \b{33}	& \b{13}	& 40		\\
\hline \textbf{Rubble 2}			& IE2				& 114.0		& \b{63.2}		& 7			& \b{3}		& \b{10}	& 26		\\
\hline \textbf{Forest 1}			& E					& \b{65.7}	& 74.4			& \b{0}		& 2			& \b{6}		& 18		\\
\hline\hline \textbf{Forest 2}		& E					& 36.8		& \b{35.7}		& N/A		& N/A		& \b{2}		& 3			\\
\hline \textbf{Forest 3}			& E					& 132.1		& \b{75.3}		& N/A		& N/A		& \b{4}		& 10		\\
\hline
\end{tabular}

	\vspace{5pt}
	Each pair of columns (MC/AC) shows the medians of the 3~metrics evaluated for the experiments. 
	Of each pair, the value in bold is better.
	Experiments \emph{Forest~2} and \emph{Forest~3} are those conducted in~\cite{zimmermann-icra2014}. 
	Both robot construction and AT algorithm changed in the meantime, so the values should not be compared to the new results.
\end{table}

\vspace{-1em}

\subsection{Results}\label{sec:exp_results}

It can be seen in \autoref{tab:at_all_results} that the \emph{Time to finish} with AC tends to be shorter or comparable to MC (and with TPV, it is even shorter, as expected).
Subjectively, the operators report a much lower level of cognitive load when driving with AC, which means they can pay more attention to exploration or other tasks.

\emph{Penalties} with AC are also mostly better or comparable to MC.
The \emph{number of flipper configuration changes} for AC is approximately 2- to 4-times higher than for MC.
However, with AC, there is no time penalty for changing flipper configurations, and it also adds no more cognitive load to the operator.

From the experiments conducted it follows that AC yields similar or even better performance than MC. Furthermore, AC allows the operator to concentrate rather on higher-level tasks while having the tedious and low-level flipper control done automatically.

\section{Robustness to Missing Exteroceptive Data}\label{sec:exp:marg}

\begin{figure}[t!]
	\centering
	\includegraphics[width=\columnwidth]{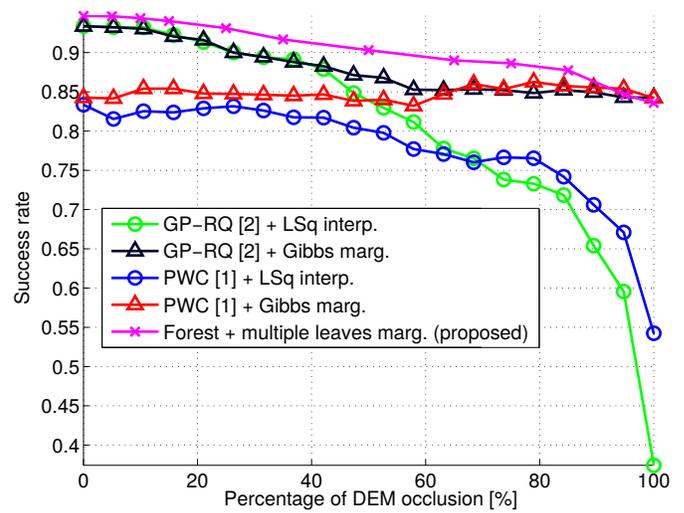}
	%
	\caption{\textbf{Robustness to DEM occlusion:} The chart shows the influence of DEM occlusion (percentage of DEM bins in which measurements are not available) on AC success rate.
	When \SI{100}{\percent} of DEM is occluded, the marginalized policies still depend on proprioceptive measurements, while LSq interpolation reconstructs only flat terrain.
	}
	\label{fig:exp:quantitative}
\end{figure}

In this experiment, we quantitatively evaluate robustness to the number of missing features for the various $Q$/QPDF representations.
The robustness is presented as the relation between success rate and the number of synthetically occluded DEM bins.

The Regression Forests first compute the marginalized QPDFs as described in \autoref{sec:Q-models}, and then choose a~configuration according to \autoref{eq:opt-Q}. The LSq interpolation/Gibbs sampling methods first interpolate or marginalize the missing data, then compute the $Q$~function on the interpolated data and choose the configuration according to \autoref{eq:best_flipper_pose}.

For this experiment, a~dataset consisting of hundreds of captured robot states (interoceptive + full exteroceptive features) is used.
The bipolar manual annotations $s_{c,\mathbf{x}}$ are assigned to all state-action combinations.

For $i=0\ldots 100$, the set ``states$_i$'' is generated from the dataset by occluding $i$~DEM bins in each of the captured states $\mathbf{x}$ (the same manual annotation $s_{c,\mathbf{x}}$ is used for all states $\mathbf{\widetilde{x}}$~generated from~$\mathbf{x}$).
To avoid combinatorial explosion, we did not try all combinations of $i$~occluded bins.
We chose to successively occlude DEM bins from the front of the robot, until $i$~bins are occluded.
Therefore, the dataset the robustness is tested on contains tens of thousands of different states.
The success rate in~\autoref{fig:exp:quantitative} is computed as $\text{success rate}(\text{states}_i)$ according to~\autoref{eq:success_ratio}.

\autoref{fig:exp:quantitative} shows superiority of marginalizing methods over LSq interpolation. Up to a~DEM occlusion level of \SI{40}{\percent}, all methods behave comparably. The reasons are two-fold: (i)~the part of the occluded DEM is far in front of the robot and there is no way to sense it from the proprioceptive measurements, (ii)~the obstacle hidden in this part of DEM is usually far enough, therefore the V-shape configuration (the one for flat terrain) is still allowed in most of the testing data. When more than \SI{40}{\percent} are hidden, success rate of the LSq interpolation method drops rapidly down towards $0.4-0.5$ (i.e. \SI{40}{\percent}-\SI{50}{\percent} of states in which the permitted configuration is selected) for both GP and PWC, while the marginalizing methods preserve high precision. 
The figure also demonstrates that the proposed Regression Forests provide better success rate than the previous methods~\cite{zimmermann-icra2014,zimmermann-icra2015}.

\section{Tactile Terrain Exploration}\label{sec:exp:TTE}

\begin{figure}[t!]
	\centering
	\includegraphics[width=\columnwidth]{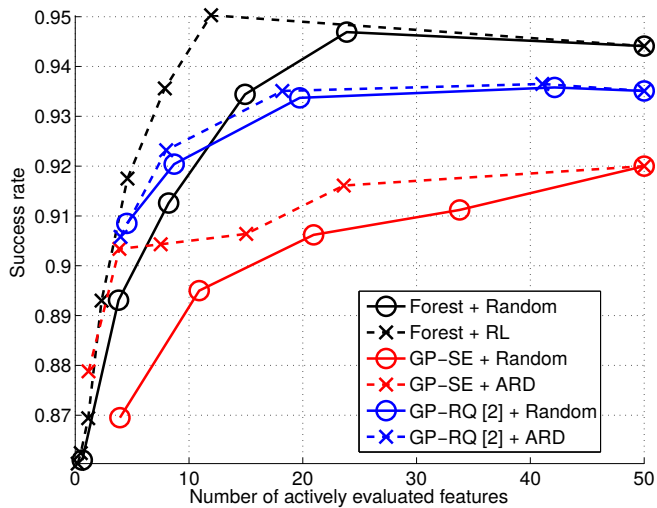}
	%
	\caption{\textbf{Comparison of TTE methods:}
		Curves on this graph show success rate (with \SI{50}{\percent} DEM bins occluded) as a function of the number of measured bin heights.
		The compared TTE strategies are described in~\autoref{sec:TTE}.
	}
	\label{fig:exp:TTE}
\end{figure}

To compare the strategies for Tactile Terrain Exploration (TTE), we evaluate them on real (test) data with the front \SI{50}{\percent} of the DEM synthetically occluded (since it is not easy to provide a~sufficient amount of real examples with naturally missing features). Active exploration is simulated by revealing the already known DEM heights.

The performance of TTE strategies can be expressed as the average number of actively measured bin heights until a safe configuration is found.
However, for this experiment, we let the exploration continue even if a safe configuration has already been found, to see how much further exploration helps.
For different QPDF models and TTE strategies, the relation between the number of measured heights and the success rate is depicted in~\autoref{fig:exp:TTE}.

An ideal QPDF model and strategy would achieve \SI{100}{\percent} success rate with a~single evaluated feature, i.e. the upper-left corner in~\autoref{fig:exp:TTE}. The closer is the curve to this corner, the better is the method. 
Results with the lowest success rate were achieved with the GP-SE method (however, the ARD strategy yields a~significant improvement). 
Better results were achieved by the GP-RQ method (for which the ARD strategy yields only small improvement compared to the Random strategy). The reason is that the RQ kernel allows for better generalization than the SE kernel. 
For less than $15$~features actively evaluated (i.e. smaller safety thresholds), the GP-RQ method achieves higher success rate than the Regression Forest method with Random strategy. 
The best method in this comparison are Regression Forests combined with the RL strategy, which achieve the best success rate.
 
\section{Conclusion}\label{sec:conclusions}
We extended the Autonomous Control algorithm~\cite{zimmermann-icra2014,zimmermann-icra2015} that increases autonomy in mobile robot control and reduces cognitive load of the operator. To deal with only partially observable terrain, missing or incorrect data, we (i) designed and experimentally verified a~more occlusion-robust QPDF model, and (ii) we exploit a~body-mounted robotic arm as an additional active sensor for Tactile Terrain Exploration. 
TTE is used in dangerous situations, where all actions have negative expected rewards.
The previous methods have to choose one of the actions, even if the best expected reward is negative.
By tactile exploration of the unobserved part of the terrain, the reward estimates get better and at least one of them should get positive if the terrain is traversable.
Several TTE strategies were proposed and experimentally evaluated. We conclude that the overall highest success rate was achieved by combining Regression Forests with the RL strategy for the arm-based exploration of missing data.

\bibliographystyle{Bibliography/IEEEtranTIE}
\bibliography{Bibliography/IEEEabrv,Bibliography/BIB_15-TIE-3764}


\if@shownames
\vspace{-2cm}
\begin{IEEEbiography}[{\includegraphics[width=1in,height=1.25in,clip,keepaspectratio]{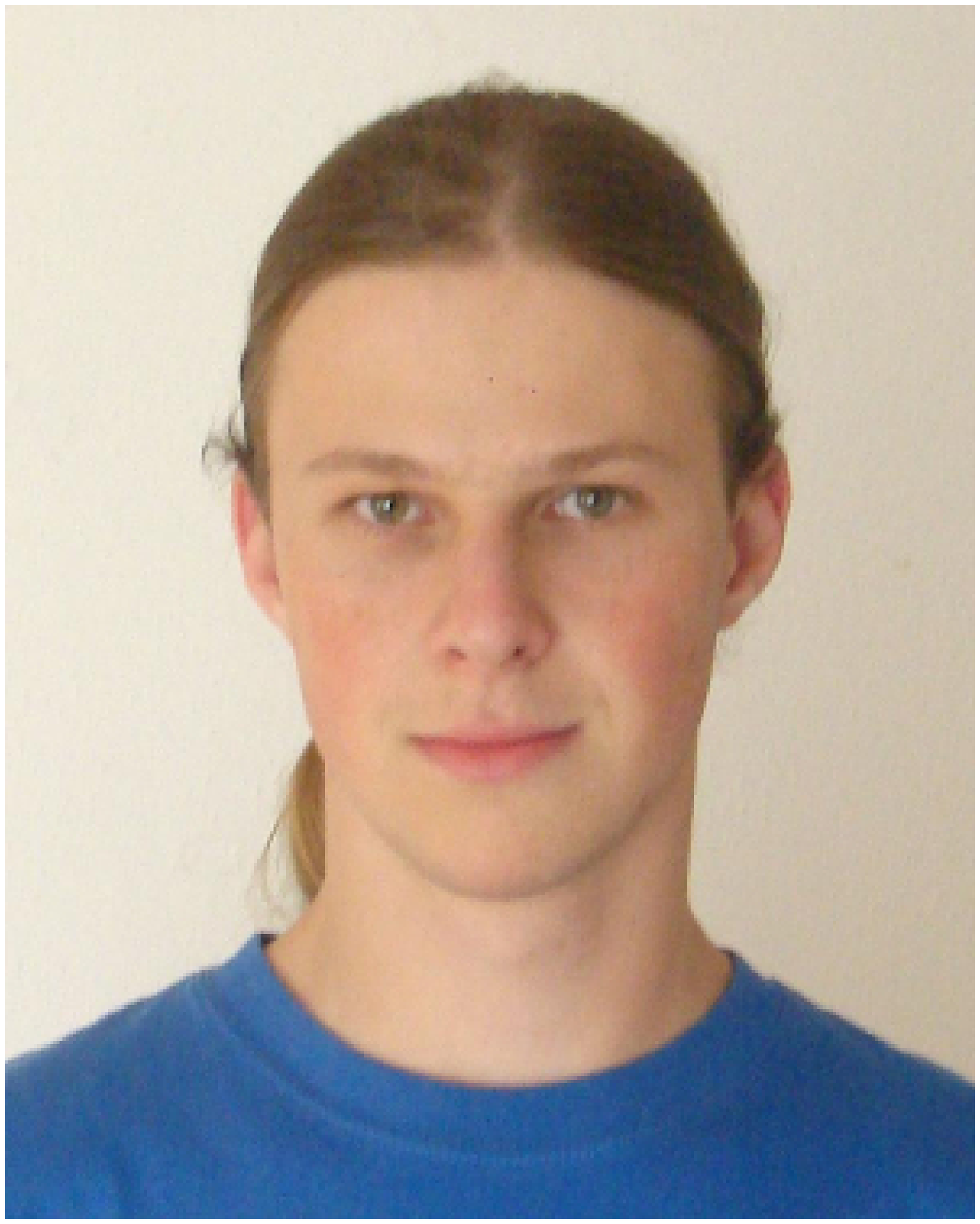}}]{Martin Pecka}
received the Mgr. (M.Sc.) degree in theoretical informatics at the Faculty of Mathematics and Physics, Charles University in Prague, Czech republic, in 2012.

He has currently been a Ph.D. student at the Department of Cybernetics, Czech Technical University in Prague (CTU), and a Research Assistant at the Czech Institute of Informatics, Robotics and Cybernetics, CTU in Prague. 
His main research interests are in the fields of machine learning, specifically reinforcement learning concerning safety of execution, and robotics.

\end{IEEEbiography}
\begin{IEEEbiography}[{\includegraphics[width=1in,height=1.25in,clip,keepaspectratio]{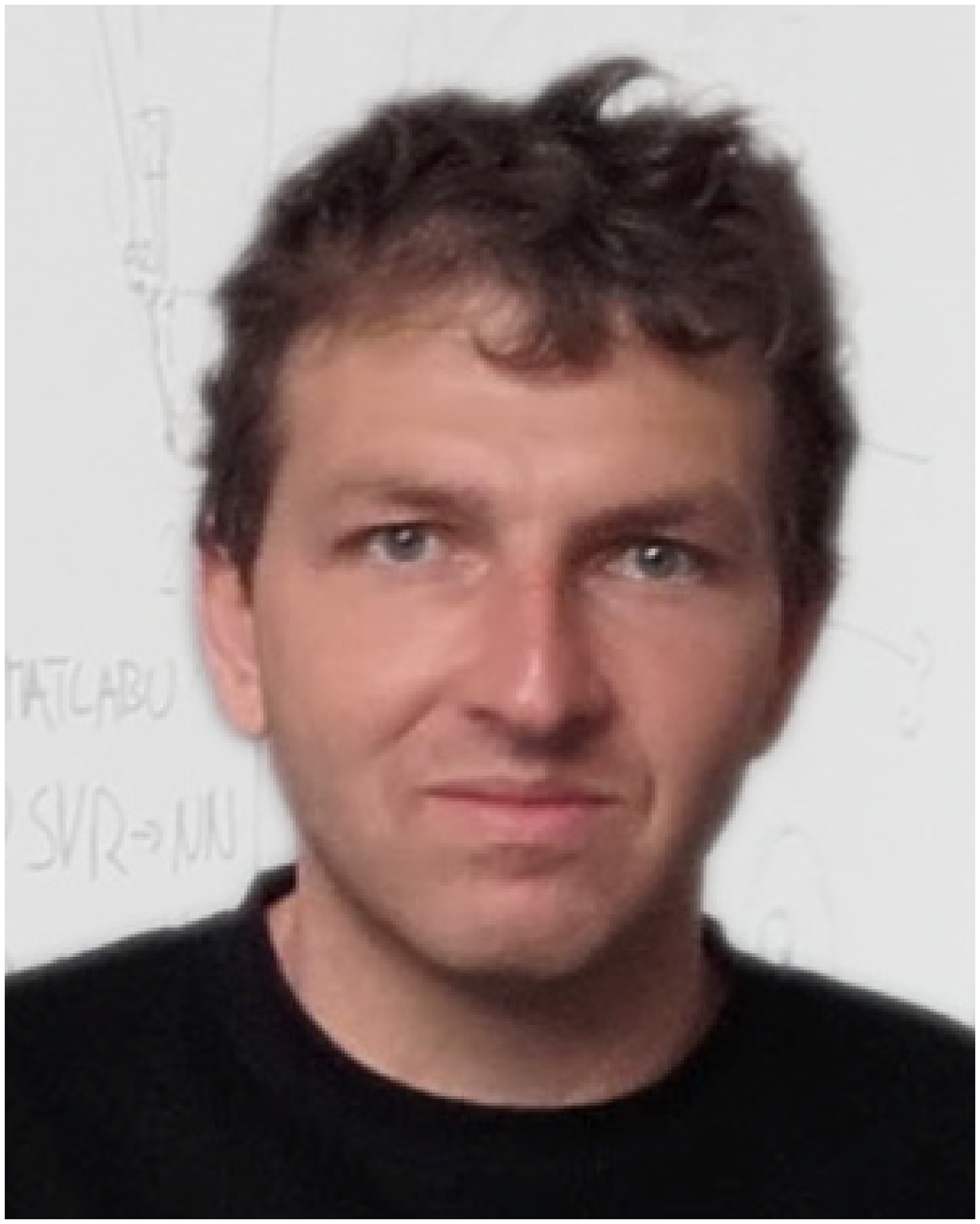}}]{Karel Zimmermann}
(M'08) received the Ph.D. degree in cybernetics from the Czech Technical University in Prague, Czech Republic, in 2008. 

He worked as Postdoctoral Researcher with the Katholieke Universiteit Leuven (2008-2009). 
Since 2009, he has been a Postdoctoral Researcher at the Czech Technical University in Prague. 
His current research interests include learnable methods for tracking, detection and robotics.

Dr. Zimmermann serves as a reviewer for major journals such as TPAMI, IJCV and conferences such as CVPR, ICCV, ICRA, IROS. 
He received the best reviewer award at CVPR 2011 and the main prize for the best PhD thesis in Czech Republic in 2008 (awarded by the Czech Society for Pattern Recognition).

\end{IEEEbiography}
\begin{IEEEbiography}[{\includegraphics[width=1in,height=1.25in,clip,keepaspectratio]{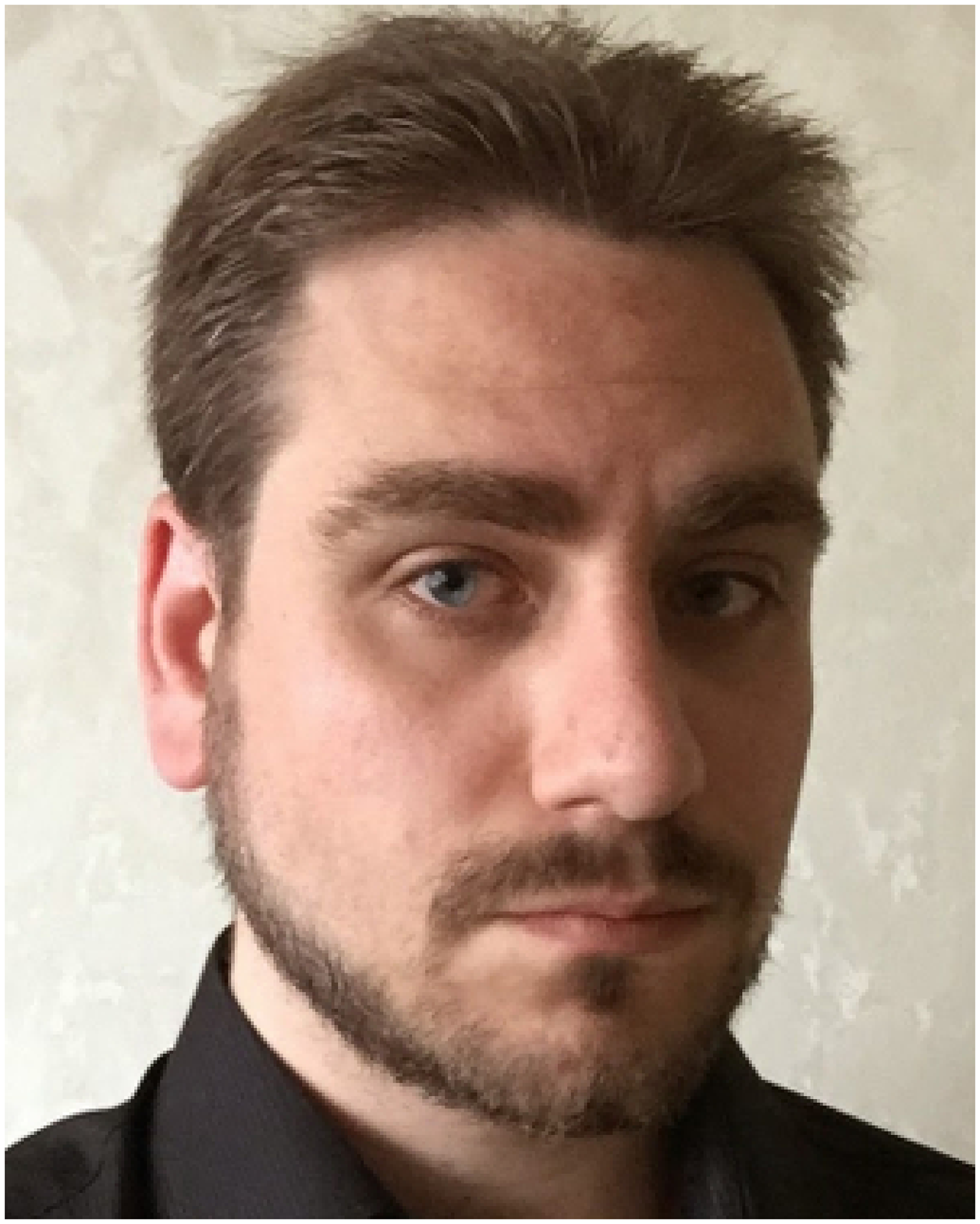}}]{Michal Reinstein}
(M'11) received the Ing. (M.Sc.) and Ph.D. degrees in engineering of aircraft information and control systems from the Faculty of Electrical Engineering, Czech Technical University in Prague (CTU), Czech republic, in 2007 and 2011, respectively. 

He is currently working as Assistant Professor at the Center for Machine Perception, Dept. of Cybernetics, CTU in Prague. His most recent research interests concern application of machine learning and data fusion to satellite imagery aiming to support deployment of large-scale robotic platforms. In the past years since 2011, the main topics of his research were: sensory-motor interaction in legged robots, multimodal data fusion for robots intended for Urban Search \& Rescue, and adaptive traversability of unknown terrain using reinforcement learning.
\end{IEEEbiography}
\begin{IEEEbiography}[{\includegraphics[width=1in,height=1.25in,clip,keepaspectratio]{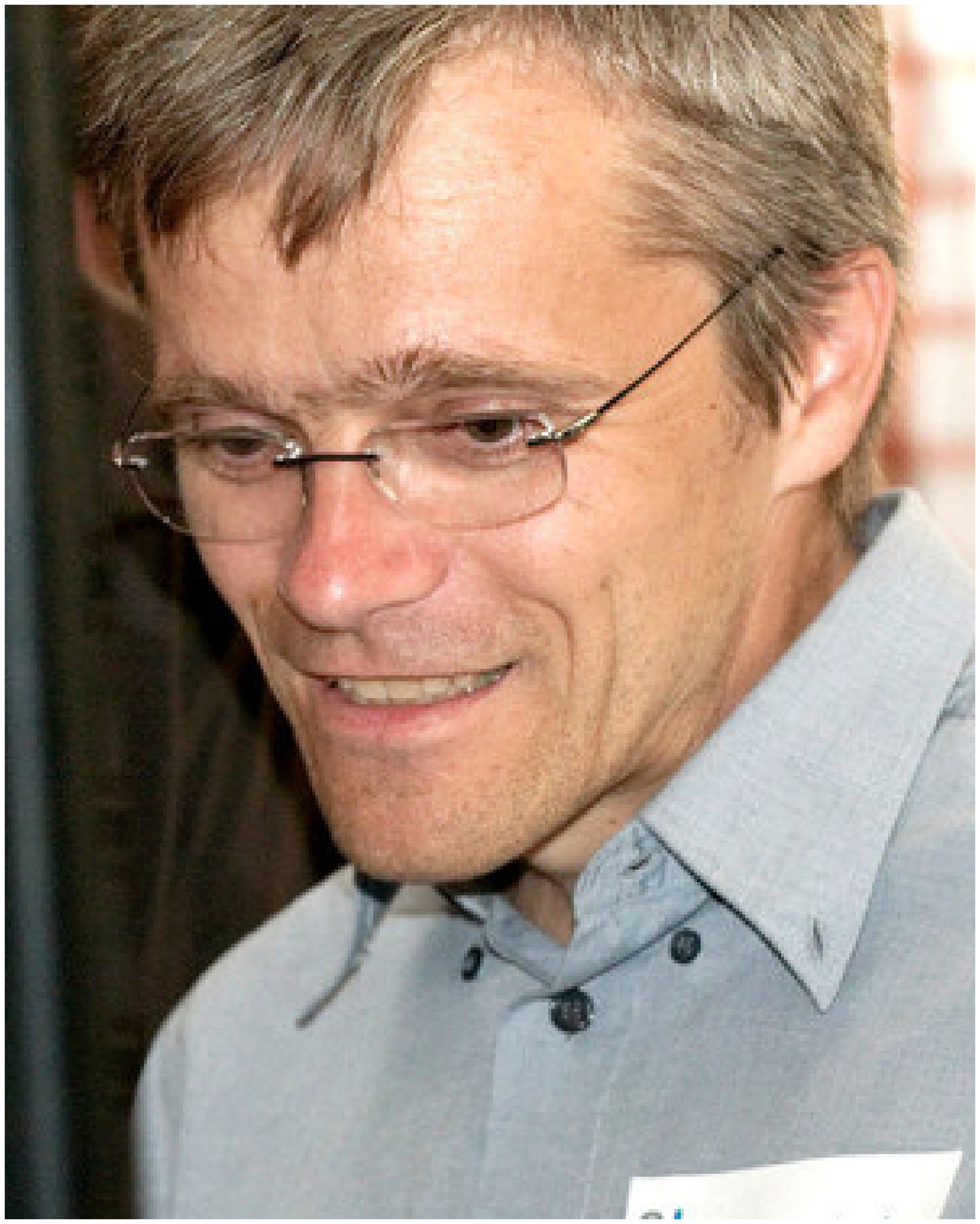}}]{Tomas Svoboda}
(M'01) received the Ph.D. degree in artificial intelligence and biocybernetics from the Czech Technical University in Prague, Czech republic, in 2000.

Later, he spent three post-doc years with the Computer Vision Group at the ETH Zurich.
Currently, he is Associate Professor and Deputy Head of the Department of Cybernetics at the Czech Technical University in Prague, the Director of EECS study programme, and he is also on board of Open Informatics programme.
He has published papers on multicamera systems, omnidirectional cameras, image based retrieval, learnable detection methods, and USAR robotics.  His current research interests include multimodal perception for autonomous systems, object detection and related applications in automotive industry.

\end{IEEEbiography}
\vspace{6cm}
\fi

\end{document}